\documentclass[preprint]{elsarticle}

\usepackage{amsmath}
\usepackage{amsfonts,amssymb}
\usepackage{bm}
\usepackage{cases}
\usepackage{subfigure} 
\usepackage{booktabs}
\usepackage{longtable}
\usepackage{url}
\usepackage{multirow}
\usepackage{listings}
\usepackage{algorithm}  
\usepackage{algpseudocode} 
\usepackage{marvosym}
\newcommand{\tabincell}[2]{\begin{tabular}{@{}#1@{}}#2\end{tabular}} 
\usepackage{graphicx} %
\usepackage{epsfig}
\usepackage{fancyhdr}
\usepackage{setspace}
\usepackage{helvet}
\usepackage{makecell}
\usepackage[strings]{underscore}

\makeatletter
\def\ps@pprintTitle{\let\@oddhead\@empty\let\@evenhead\@empty\let\@oddfoot\@empty\let\@evenfoot\@empty}
\makeatother

\begin{document}
\begin{frontmatter}
\title{A Dynamic Aggregation Strategy Enhanced Efficient Global Optimization Algorithm for Solving High-Dimensional Turbomachinery Design Problems}

\author[1]{Qineng Wang%
}
\ead{zhet1997@stu.xjtu.edu.cn}

\author[1]{Zhendong Guo%
}
\ead{guozhendong@xjtu.edu.cn}

\author[2]{Yun Chen%
}
\ead{saeri.chen@qq.com}

\author[2]{Guangjian Ma%
}
\ead{maguangjian0401@163.com}

\author[1]{Liming Song\corref{cor1}%
}
\ead{songlm@xjtu.edu.cn}

\author[1]{Jun Li%
}
\ead{junli@mail.xjtu.edu.cn}

\cortext[cor1]{Corresponding author}

\affiliation[1]{organization={Institute of Turbomachinery, Xi'an Jiaotong University},
addressline={No.28, West Xianning Road},
postcode={710049},
city={Xi'an},
country={China}}

\affiliation[2]{organization={AVIC Shenyang Engine Design Institute},
addressline={Fangjialan Road, Wanlian Street},
postcode={110066},
city={Shenyang},
country={China}}

\begin{abstract}
In order to solve the high-dimensional ($d \geq 30$) expensive black-box problems within budget, an efficient global optimization (EGO) algorithm with a dynamic aggregation strategy is proposed, labeled as DA-EGO. 
Specifically, the DA-EGO decomposes the original high-dimensional design space into a set of low-dimensional subspaces for efficient surrogate-based optimization search, and the optimal solutions of subspaces are combined as an elite point for the global search. 
Most importantly, the subspaces are not fixed. 
Instead, the subspace variables are updated in each iteration, according to the variable interaction analyses in the sub- and full-spaces. 
The perturbation method and the analysis of variance are used to detect variable interactions. 
To further accelerate the optimization progress, the searching ranges of subspaces are also adaptively adjusted according to the analyses of subspace optimization results of the previous iteration. 
Tests on 21 benchmark instances, comprising seven functions at 30, 60, and 90 dimensions, show that DA-EGO is effective on separable and partially separable problems under a budget of 1500 function evaluations. Its advantage is case-dependent: on the non-separable shifted Rosenbrock function, GSGA performs better at 60 and 90 dimensions, while the 30-dimensional results are statistically comparable to IKAEA and GSGA. 
Moreover, the advantage of DA-EGO is also seen in the aerodynamic optimization of a transonic rotor blade with 28 variables as well as the compressor stage optimization with 60 variables. 
With the above, the effectiveness of the proposed DA-EGO has been well demonstrated.
\end{abstract}


\end{frontmatter}

\section{Introduction}
\par
The appearance of simulation-based engineering optimization has greatly improved the efficiency and quality of engineering design, so it is widely used in various engineering fields. 
However, when the engineering problem to be solved is characterized by high dimensionality or expensive, the difficulty of optimization will be greatly increased~\citep{chenMeasuringCurseDimensionality2015}. 
These two characteristics are ubiquitous in the field of engineering design~\citep{shanSurveyModelingOptimization2010}. 
Taking the aerodynamic shape design optimization as an example~\citep{tangAdaptiveDynamicSurrogateassisted2022,tangHierarchicalVariableFidelity2022a}, the high dimensionality characteristic is reflected in the complex curved shape and a large number of design details of aerodynamic components (such as cascades and wings), which require tens or even hundreds of design variables to accurately describe and adjust. 
The aerodynamic performance of the components that are designed needs to be obtained through wind tunnel experiments or computational fluid dynamics (CFD) simulations. 
High-fidelity CFD simulations are very time-consuming, individual samples can take hours or even days to evaluate. 
According to different research fields, the specific definitions of high dimensionality and expensive are also different in literature. 
Following the definition in ~\cite{zhanFastKrigingAssistedEvolutionary2021}, the high dimensionality means that the number of variables varies from 30 to 100, and expensive refers to that the samples available for the whole optimization process are limited to thousands.
\par
The optimization process posed a significant computational challenge resulting from the presence of high-dimensional or computationally expensive characteristics~\cite{tangAdaptiveDynamicSurrogateassisted2022}. 
In tackling high-dimensional problems with dimensions greater than 30, conventional population-based algorithms require millions of samples~\cite{okulewiczSelfAdaptingParticleSwarm2022}. 
Consequently, such optimization processes often become computationally intensive. 
Various optimization algorithms developed by researchers have been applied to mitigate this challenge. However, these methods are largely ineffective in solving high dimension and expensive black-box (HEB) problems.
\par
In the optimization of high-dimensional engineering problems, the increase of the number of design variables will lead to the exponential expansion of the design space and the sharp increase of the complexity of the relationship between variables, so that the number of samples needed to complete the optimized search or build the surrogate will also increase exponentially. 
This difficulty caused by high dimensions is also called ``curse of dimensionality".
To solve the high-dimensional problems, the optimization algorithm based on decomposition (also known as divide and conquer) is the most widely used~\citep{sunRecursiveDecompositionMethod2018,meiCompetitiveDivideandConquerAlgorithm2016}. 
It decomposes the high-dimensional original problem into several smaller and simpler subproblems for optimization. 
How to decompose the problem is the key of this kind of algorithm. 
When the decomposition algorithm was put forward, the decomposition scheme was always completely independent of the original problem~\cite{potterCooperativeCoevolutionaryApproach1994,yangLargeScaleEvolutionary2008}, that is, the variables were divided into multiple combinations in a fixed or random way, without considering the characteristics of the problem. 
Subsequent studies have proved that such a simple ``manual decomposition" strategy is unable to deal with problems with complex variable interaction~\citep{omidvarCooperativeCoEvolutionDifferential2014,meselhiDecompositionApproachLargescale2022}. 
Then, the variable interaction-based decomposition (VID) method has received more widespread attention~\cite{omidvarCooperativeCoevolutionDelta2010,chenLargeScaleGlobalOptimization2010,omidvarCooperativeCoEvolutionDifferential2014,liuDynamicMultiplePopulations2018}, the VID method divided variables into groups according to their interaction relationship between each other. 
However, it is not suitable for expensive engineering optimization, because it requires numerous samples to obtain the interaction structure of the problem.
\par
In the optimization of expensive engineering problems, the main difficulty is that the increase of single sample cost inevitably leads to the decrease of the total number of available samples. 
For this reason, researchers developed the surrogate-based optimization (SBO) algorithms to solve expensive problems~\cite{forresterOptimizationUsingSurrogate2006,qianBayesianHierarchicalModeling2008}. 
The SBO method builds an approximate model based on evaluated samples, which significantly reduces the number of samples required by replacing expensive sample evaluation with the explicit surrogate model. 
A SBO method usually has two key components: the establishment of surrogate and the search with an acquisitive function.
The former summarizes the knowledge in the design space, while the latter uses the knowledge to explore efficiently. 
However, due to the curse of dimensionality, common surrogate technologies cannot effectively extract the knowledge from the high-dimensional design space, which is also reflected in the decline of the accuracy of the surrogate. 
As the efficiency of the SBO algorithm largely depends on the accuracy of the established surrogate, it can only solve problems with less than 15 dimensions efficiently.
\par
In order to efficiently solve the HEB problem, researchers try to combine the surrogate into the decomposition method. 
The Nash-EGO algorithm~\citep{xuNashGameBased2018} directly combines efficient global optimization (EGO) with a manual decomposition strategy. 
The SEE algorithm~\citep{yangTurningHighDimensionalOptimization2018}  uses a novel surrogate in subtask optimization. 
In CBO-HGST algorithm~\citep{jiangCooperativeBayesianOptimization2022}, sub-optimization based on transfer Gaussian process regression is studied. 
The efficiency improvements of the above three algorithms are only reflected in the subspace, the simple manual decomposition strategy is still used globally. 
Besides, in the OMID algorithm~\citep{hajikolaeiHighDimensionalModel2014}, the global surrogate is first established, based on which the interaction relationship of all variables is estimated. 
However, the algorithm based on the high-dimensional surrogate is not immune to the curse of dimensionality, so it is only applicable to the problem with less than 30 dimensions. 
The above algorithms only use the surrogate in a single aspect to reduce costs to a certain extent within the original framework, but they still cannot solve high-dimensional problems efficiently.
\par
In view of the above situation, a new dynamic aggregate efficient global optimization algorithm (DA-EGO) is proposed in this paper, which is suitable for HEB problems.  
Similarly, the high-dimensional problem is also decomposed into multiple low-dimensional subproblems in the DA-EGO. 
The basic idea of the DA-EGO algorithm is to use the surrogate model to conclude the knowledge from the subspaces. Then, this ``local" knowledge is aggregated together into ``global" knowledge of the original design space and used to generate the decomposition scheme in the next cycle.
The DA-EGO algorithm proposed in this paper is improved from the following aspects in detail: 
(\romannumeral1) 
Unlike the traditional VID algorithm, which completes the decomposition of space and the construction of subproblems at one time, the DA-EGO algorithm has multiple cycle of iteration, and in each cycle, the space decomposition method and the setting of subproblems will gradually change from rough to detailed.
(\romannumeral2) 
In the DA-EGO algorithm, subtasks can feed back the decomposition strategy. 
Knowledge mining after subspace optimization can provide more information about the problem. 
On the one hand, it can verify the accuracy of existing information, and on the other hand, it can provide new information to help the construction of subsequent subtasks.
(\romannumeral3) 
In DA-EGO, a space reduction mechanism is added, and more knowledge about design space is used to guide the optimization of subspace. 
\par
The remainder of the paper is organized as follows. 
In Section \uppercase\expandafter{\romannumeral2}, the related works of DA-EGO is introduced.
And then, the details of the proposed DA-EGO algorithm are illustrated in Section \uppercase\expandafter{\romannumeral3}.
After that, the DA-EGO algorithm is tested on 18 numerical benchmark functions in Section \uppercase\expandafter{\romannumeral4}.
In Section \uppercase\expandafter{\romannumeral5}, the DA-EGO algorithm is used to optimize the 28-dimensional Rotor 37 transonic compressor cascade, and the 60-dimensional typical multistage compressor. 
The correctness and effectiveness of the proposed method are well verified.
And finally, the conclusions are summarized in Section \uppercase\expandafter{\romannumeral6}.
\section{Related Work}
\subsection{Efficient global optimization}
\par
The efficient global optimization algorithm is widely used in engineering optimization for its high efficiency~\citep{songResearchMetamodelBasedGlobal2016}. 
The EGO algorithm achieved a great balance between global exploitation and local exploration with two core components, 
the kriging surrogate, and EI acquisitive function.%
\subsubsection{Kriging surrogate}
\par
Kriging is a popular surrogate technique~\citep{jonesEfficientGlobalOptimization}. 
The kriging prediction $Y_{KG}$ at unknown site $x$ is built as a trend function $f(x)$ plus a normal random process $Z(x)$ as:
\begin{equation}
{Y_{KG}}({\bf{x}}) = f({\bf{x}}) + Z({\bf{x}})
\end{equation}
where, $f(x)$ is usually a constant, linear or quadratic polynomial, and the constant is most widely used; $Z(x)$ describes the local features of $Y$ around the $n$ sample points $X = \{ x^{(1)},\cdots,x^{(n)}\}$, which has zero mean and a co-variance function as:
\begin{small}
	\begin{equation}
	{\mathop{\rm cov}} \left[ {Z({\bf{x}}),Z({{\bf{x}}^{(i)}})} \right] = {\sigma ^2}\exp \left( { - \sum\limits_{h = 1}^d {{\theta _h}{{\left\| {{x_h} - x_h^{(i)}} \right\|}^2}} } \right)
	\end{equation}
\end{small}
\par 
The function prediction and related mean squared error (MSE) at an unknown point $\bf{x}$ can be expressed as:
\begin{small}
	\begin{equation}\label{kriging}
	\begin{split}
	&{{\hat y}_{KG}}({\bf{x}}) = \hat \mu  + {{\bf{r}}^T}{{\bf{R}}^{ - 1}}({\bf{y}} - {\bf{1}}\hat \mu )\\
	&{s_{KG}}({\bf{x}}) = 
	{\sigma ^2} \{1 - {\bf{r}}^T ({\bf{x}}) {\bf{R}}^{-1} {\bf{r}}({\bf{x}}) + \\
	&( 1 - {\bf{1}}^T  {\bf{R}}^{ -1} {\bf{r}}^T ({\bf{x}}))
	{{( {{{\bf{1}}^T}{{\bf{R}}^{-1}}{\bf{1}}})}^{ - 1}}
	( 1 - {\bf{l}}^T  {\bf{R}}^{-1}  {\bf{r}}^T ({\bf{x}}) )^T \}\\
	\end{split}
	\end{equation}
\end{small}
where $\mu$ is the regression constant as 
$\mu=\left(
\mathbf{1}^{T} \mathbf{R}^{-1} \mathbf{1}
\right)^{-1} 
\mathbf{1}^{T} \mathbf{R}^{-1} \mathbf{y}_{S}$;
$\mathbf{R}$ is the correlation matrix $\mathbf{R}:=\left(R\left(\mathbf{x}^{(i)}, \mathbf{x}^{(j)}\right)\right)_{i, j} \in \mathbb{R}^{n \times n}$, and $\mathbf{r}$ is the correlation vector
$\mathbf{r}:=\left(R\left(\mathbf{x}^{(i)}, \mathbf{x}\right)\right)_{i} \in \mathbb{R}^{n}$.
\subsubsection{ Maximum expectation improvement}
\par 
Once the kriging surrogate is constructed, the location with the maximum EI value can be obtained by using an arbitrary optimizer. 
The EI function at location $x$ can be defined as follows:  
\begin{equation}
  \begin{array}{c}
  EI({{\bf{x}}}) = \left( {{f_{\min }} - {\hat y}({{\bf{x}}_{sub,j}})} \right)\Phi (u) + s({{\bf{x}}_{sub,j}})\phi (u)\\
  u = {{\left( {{f_{\min }} - {{\hat y}_{}}({{\bf{x}}_{sub,j}})} \right)} \mathord{\left/
   {\vphantom {{\left( {{f_{\min }} - {{\hat y}_{}}({{\bf{x}}_{sub,j}})} \right)} {s({{\bf{x}}_{sub,j}})}}} \right.
   \kern-\nulldelimiterspace} {s_{}({{\bf{x}}_{sub,j}})}}
  \end{array}
\end{equation}
where $\Phi$ is the normal cumulative distribution function, and $\phi$ is the normal probability density function.
function. 
\par 
The conventional SBO algorithms, represented by EGO, are typically limited in their applicability to problems with up to 15 dimensions. Therefore, one of the aims of this study is to extend these sample-efficient methods to higher-dimensional problems.
\subsection{Decomposition-based optimization algorithms}
In real-world engineering problems, not all variables strongly interact with each other. 
For example, consider the following six-dimensional problem where interactions occur between $x_{2},x_{3},x_{4}$ and $x_{5},x_{6}$.
\begin{equation}\label{6d_example}
f({\bf{x}}): = x_1^2 + {\left( {{x_2} - {x_3}} \right)^2} + {\left( {{x_3} - {x_4}} \right)^2} + {\left( {{x_5} + {x_6}} \right)^2}
\end{equation}
The original problem can be divided into multiple simpler subproblems that do not interact or weakly interacted with each other:
\begin{equation}\label{6d_decompose}
\begin{array}{l}
\arg {\min _{\left( {{x_1}, \cdots ,{x_6}} \right)}}f\left( {{x_1}, \cdots ,{x_6}} \right)\\
  = \left( {\arg {{\min }_{{x_1}}}f\left( {{x_1}} \right), \cdots ,\arg {{\min }_{({x_5},{x_6})}}f\left( {{x_5},{x_6}} \right)} \right)
\end{array}
\end{equation}
Here, the six variables are divided into three groups, 
and the optimal location of the original six-dimensional problem can be obtained indirectly by solving three sub-problems with lower dimensionality.   
By utilizing a decomposition strategy, the issue of dimensionality is greatly reduced since the cost of the problem increases linearly with the number of variables rather than exponentially. 
\par 
As the interaction between sub-problems weakens, the optimization solutions of sub-problems tend to approach the true optimal solution, but obtaining the relationships among variables require higher costs.
Considering the above trade-off, decomposition strategies can be divided into manual decomposition and variable interaction based decomposition (VID). 
The manual decomposition approach, which indiscriminately separates variables into multiple groups regardless of their interaction relationships, is limited in its effectiveness and is only suitable for fully separable problems. %
Therefore, this paper mainly focuses on the VID method.
\par
\begin{figure}[ht]
\begin{center}
\includegraphics[scale=0.6, trim = 0 0 0 0]{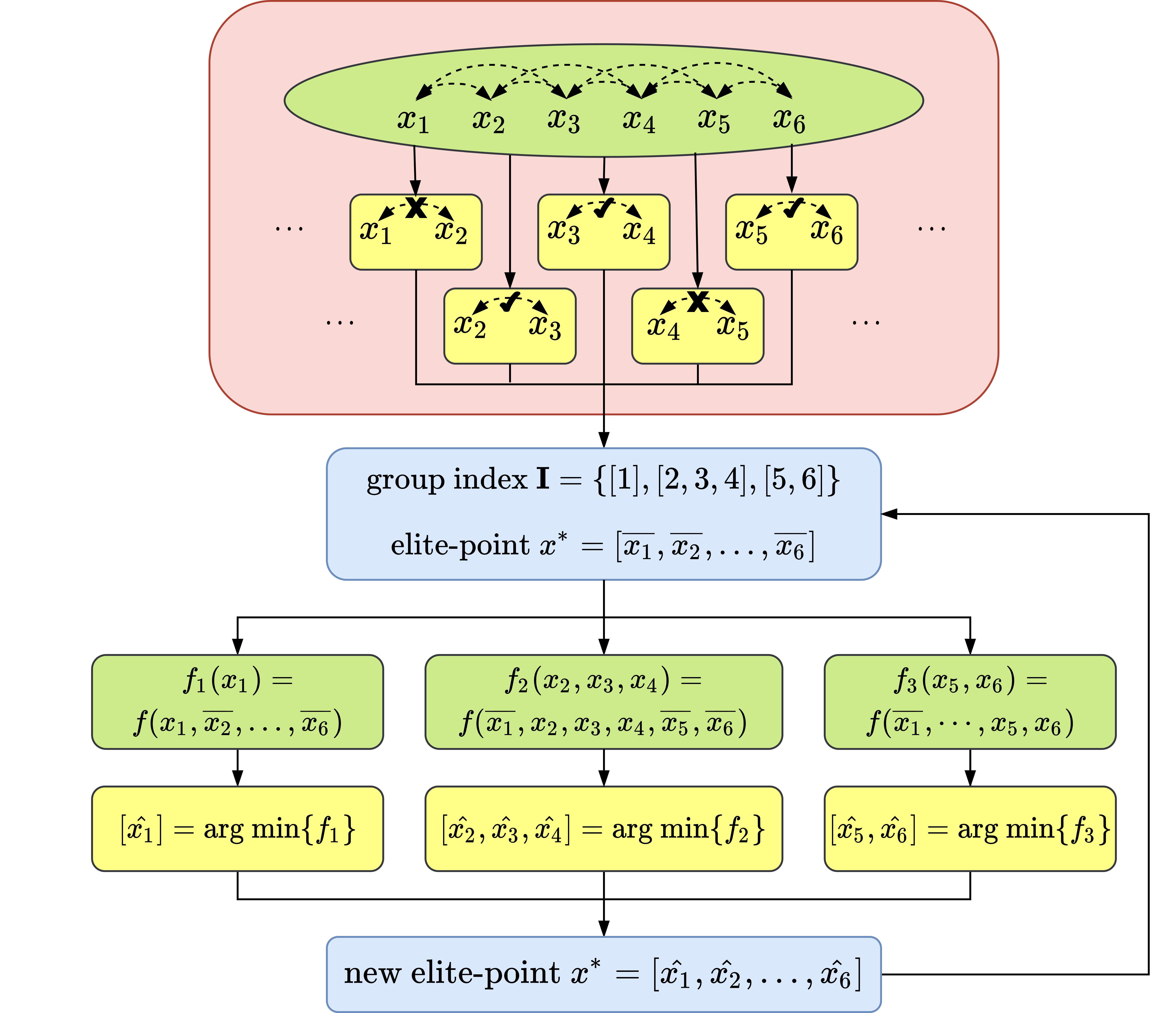}
\end{center}
\caption{Framework of the variable interaction based decomposition algorithm}   
\label{DCbased}
\end{figure}

Variable interaction based decomposition (VID) algorithms are the most popular method for addressing high-dimensional optimization problems. 
The VID algorithms' vital strategy is to identify variable interactions and leverage the group index $\mathbf{I}$ and elite-point $x^*$ to express specific decomposition arrangements~\citep{yangLargeScaleEvolutionary2008}.
As shown in Fig.\ref{DCbased}, the VID algorithm is used on Equation \ref{6d_example}. 
The first step of decomposition is detecting the interaction of each variable pairs, 
which requires $D(D-1)/2$ independent sets of samples.
Base on the variable interaction information, the group index 
${\bf{I}}=\{I_{1},I_{2},\cdots,I_{r}\}=\{[x_1],[x_2,x_3,x_4],[x_5,x_6]\}$ 
is obtained by assigning interact variables into the same group. 
Then, select the current optimal evaluated samples as the elite-point.
Sub-tasks are generated with elite-point $x^{*}$ and group index $\bf{I}$ as the Equation (\ref{gen_sub}).\\
\begin{equation}\label{gen_sub}
\begin{array}{c}
X_{\text{local}} = [x_{1}^{l},x_{2}^{l},\cdots,x_{||I_{j}||}^{l}],
X_{\text{global}} = [x_{1}^{g},x_{2}^{g},\cdots,x_{D}^{g}]\\
f_{j}^{\text{sub}}(X_{\text{local}}) = f(G(X_{\text{local}},I_{j},x^{*}))= f(X_{\text{global}})\\
x_{k}^{g} = 
\left\{
  \begin{array}{l}
  x_{k}^{l}  \text{  if   }  x_{k} \in I_{j}\\
  x_{k}^{*}  \text{  if   }  x_{k} \notin I_{j}\\
  \end{array}
\right.
\end{array}
\end{equation}
where for samples in a $||I_{j}||$ dimensional subspace, the values of remaining $D-||I_{j}||$ variables are keep fixed with the elite-point.
\par
Presently, the greatest challenge of implementing VID algorithms for expensive problems is the high cost of sampling for interaction structure detection. 
And the proposed DA-EGO algorithm introduced in this paper establish a new framework that significantly reduces the number of samples needed for variable interaction detection.
\subsection{Polynomial chaos expansion based sensitivity analysis}
\par
Polynomial chaos expansion (PCE)~\citep{sudretGlobalSensitivityAnalysis2008,wienerHomogeneousChaos1938} is a well-known surrogate technique that represents random variables in terms of polynomial functions based on other random variables. 
It serves as a surrogate model with fitting regression capabilities, especially when working with uniformly distributed random variables. Moreover, an essential feature of PCE is that its coefficients contain critical information about the surrogate model's sensitivity analysis. 
This information can be exploited to compute global sensitivity indices effectively and at significantly lower costs.
\par 
Formally, the function prediction of PCE can be expressed as:
\begin{equation}
f(x) \approx {{\cal M}^{(PCE)}}(x) = 
\sum\limits_{{\bf{\alpha }} \in {\cal A}} {a_{\bf{\alpha }}} {\psi _{\bf{\alpha }}}(x)
\end{equation}
where, ${\psi _{\bf{\alpha }}}(x)$ is a sequence of polynomials; 
$\alpha$ is the multi-index of the multivariate polynomial ${\psi _{\bf{\alpha }}}(x)$,
${\bf{\alpha }} = \{ {{\alpha _1}, \cdots ,{\alpha _D}} \}$;
$D$ is the dimension of input variables.
Multivariate polynomial ${\psi _{\bf{\alpha }}}(x)$ is the product of multiple orthogonal single-variable polynomials $\psi _{{\alpha _i}}^{(i)}\left( {{X_i}} \right)$, 
as ${\psi _{\bf{\alpha }}}(x) = \prod\limits_{i = 1}^D {\psi _{{\alpha _i}}^{(i)}} (x_i)$.
\par 
In our study, the PCE will be built with uniformly distributed variables. Correspondingly, the Legendre polynomial is selected as the orthogonal base when building PCE as expressed in Equation \ref{legendre}, where, ${\delta _{ij}}$ is 1 if $i = j$ and 0 if $i \neq j$.
\begin{equation}\label{legendre}
{\left\langle {\psi _i^{(k)},\psi _j^{(k)}} \right\rangle _k} = \int_{{{\cal D}_k}} {{\psi _i}} (x){\psi _j}(x){f_{{X_k}}}(x)dx = {\delta _{ij}}
\end{equation}
\par 
Set $\mathbf{v} \stackrel{\text { def }}{=}\left\{i_1, \ldots, i_k\right\} \subset\{1, \ldots, M\}$ and denoting by $x_v$ the subvector of
$x$ is obtained by extracting the components labeled by the indices in $v$. 
Extract all relevant terms of $v$ from the PCE expression as follow:
\begin{equation}
f_{\mathbf{v}}\left(x_{\mathbf{v}}\right)=
\sum_{\boldsymbol{\alpha} \in \mathcal{A}_{\mathbf{v}}} \widehat{y}_{\boldsymbol{\alpha}} \Psi_{\boldsymbol{\alpha}}(\boldsymbol{x})
\end{equation}
The associated sensitivity indices are just the ratio of the above two quantities as shown in Equation \ref{varPCE}.
\begin{equation}\label{varPCE}
  \begin{aligned}
  \operatorname{Var}\left[Y_{\mathcal{A}}\right] &=\sum_{\substack{\alpha \in \mathcal{A} \\
  \alpha \neq 0}} \widehat{y}_{\boldsymbol{\alpha}}^2, \\
  \operatorname{Var}\left[f_{\mathbf{v}}\left(\boldsymbol{x}_{\mathbf{v}}\right)\right] &=\sum_{\substack{\alpha \in \mathcal{A}_{\mathbf{v}} \\
  \boldsymbol{\alpha} \neq \mathbf{0}}} \widehat{y}_{\boldsymbol{\alpha}}^2,
  \end{aligned}
\end{equation}
The PCE coefficients are calculated using Orthogonal Matching Pursuit (OMP)~\citep{wangGeneralizedOrthogonalMatchingPursuit2012}. The construction of the polynomial basis, coefficient calculation, and PCE-based sensitivity analysis use the MATLAB library UQLab~\citep{marelliUQLabFramework2014}. More details on PCE can be found in~\citep{xiuNumericalMethodsStochastic2010}.
\section{Proposed Method}
\par
As mentioned in the introduction, the DA-EGO algorithm is proposed to solve the wildly existing HEB problems in the field of engineering design. 
Figure \ref{flowchart} shows the overall framework of the DA-EGO algorithm, which can be mainly divided into three parts: (\romannumeral1) initialization, (\romannumeral2) decomposition strategy of high-dimensional space, (\romannumeral3) sub-problem optimization and knowledge mining.
After the initialization of the problem is completed, 
the decomposition strategy part generates multiple simpler sub-tasks based on existing knowledge of the whole design space. 
Then, each sub-task is processed in the sub-problem optimization and knowledge mining part. 
After sub-optimization processing, outcomes in the subspace  such as samples, interactive effects, and search boundaries are obtained and fed back to the decomposition strategy part.
These two parts will proceed in turn until the optimization stop condition is met. 
\par
Compare with conventional VID algorithms, the DA-EGO simplifies the highly challenging task of detecting the interactions among all variables at once into multiple processes of knowledge mining at low-dimensional subspaces and aggregating them.
Such an approach effectively mitigates the curse of dimensionality in high-dimensional problems, thereby enhancing the efficiency of the solution process.
In the following sections, the DA-EGO algorithm will be introduced in detail.
\begin{figure}[htbp]
  \begin{center}
  \includegraphics[scale=0.5, trim = 0 0 0 0]{DA-EGO_flowchart.png}
  \end{center}
  \caption{The framework of the proposed DA-EGO algorithm.}   
  \label{flowchart}
\end{figure}
\subsection{ Initialization}
In the initialization part, samples, interactions, and search boundaries need to be initialized in turn. 
For the initial sampling, it uses the Latin hypercube sampling (LHS) method to uniformly generate 50 initial samples in the design space and evaluate their values. 
In addition, at the beginning of optimization, there is no interaction between all variables by default, and the search scope is the entire design space, which means the search range to each variable is $[0,1]$ in the beginning. 
\subsection{High-dimensional problem decomposition} 
\par
The role of the decomposition strategy part is to generate sub-tasks based on the existing knowledge of the whole design space. 
Its work can be divided into two steps:
(\romannumeral1) Aggregate the feedback knowledge from each sub-tasks.
(\romannumeral2) Design the decomposition plan based on the integrated existing knowledge accumulated in current and previous cycles.
\par
\begin{figure}[htbp]
  \begin{center}
  \includegraphics[scale=0.55, trim = 0 0 0 0]{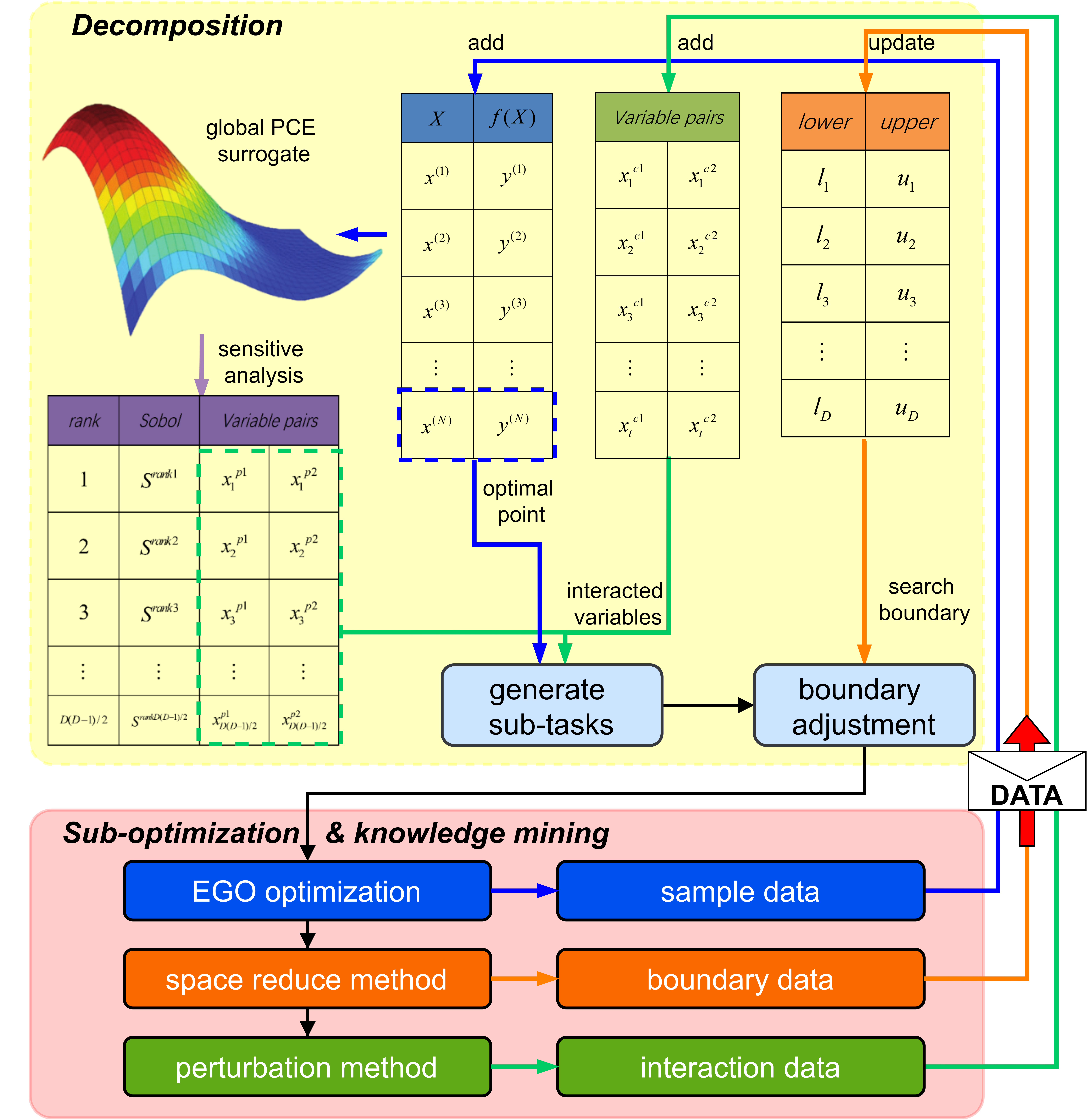}
  \end{center}
  \caption{The data flow of the proposed DA-EGO algorithm.}   
  \label{dataflow}
\end{figure}

The process of knowledge aggregation is a key component of the DA-EGO algorithm. 
Prior to discussing the aggregation methods, it is necessary to specify the information that is returned from the sub-tasks to the decomposition part.
Figure \ref{dataflow} shows the data flow diagram of the DA-EGO algorithm. 
The three kinds of knowledge reserve are respectively represented by the three tables at the top of Fig. \ref{dataflow}. 
The feedback provided by the sub-tasks includes samples, interaction relationships, and search ranges within the subspace, while the knowledge required to generate new sub-tasks includes variable interaction, optimal points, and search ranges within the global design space. 
Thus, knowledge aggregation is necessary to integrate this information together.
\par
As described in Section 2.2, the elite-point is a crucial part of decomposing a high-dimensional problem.
Here, the current best sample among all the evaluated samples will be selected as the new elite-point. 
And the aggregating of the decomposition scheme and the search scope is more complex, which will be described in detail below.
\subsubsection{Decomposition scheme aggregation}
The decomposition scheme determines which variables will be placed in the same subproblem, it is based on two sources of data in the DA-EGO. 
The first one is the pair of variables accumulated in the variable interaction table, which is denoted as ${P_C} = (x_1^{c1}x_1^{c2},x_2^{c1}x_2^{c2},\cdots,x_t^{c1}x_t^{c2})$; 
The second one is the estimate of the sensitivity values between the two variables $S_{ij}^{{\rm{PCE}}}(1 \le i < j \le D)$, which is obtained using the sensitivity analysis method based on the chaotic polynomial model (PCE) described in Section 2.2. 
Variable relationships are classified as confirmed interactions, suspected interactions, and non-interactions (which may include weak interactions). The $t$ confirmed pairs are stored in $P_C$. Among the remaining $D(D-1)/2-t$ pairs, suspected interactions are selected by sorting the PCE-based Sobol sensitivity estimates in descending order; the resulting ordered set is denoted by $P_S$. Suspected pairs are subsequently checked using the low-dimensional surrogate-based perturbation analysis. Here, $D$ is the dimension of the original problem. 
The green dotted box in Fig.\ref{dataflow} is the set of variable pairs ${P_S}$. 
It is worth noting that due to the existence of the curse of dimensionality, the accuracy of the high-dimensional PCE surrogate is limited, which will lead to the error of interactive variable pair selection. 
After the  ${P_C}$ and ${P_S}$ are obtained, the decomposition of the original problem is completed based on the following steps to obtain the final variable decomposition scheme.
The pseudo-code of the decomposition scheme update part is shown in Algo.1.
\begin{algorithm}  
  \caption{Aggregate the decomposition scheme}  
  \begin{algorithmic}[1]
  \Require 
  \Statex the dataset of sorted variable pairs($P_{S}$);
  the dataset of accumulated interacted variable pairs($P_{C}$);
  \Ensure 
  \Statex  the final decomposition scheme(${\bf{I}}$);
  \For{$i= 1 \to D$} 
  \State${I_{i} \gets x_{i}}$
  \EndFor

  \For{$k=1 \to t$}
  \State $[x_k^{c1},x_k^{c2}]\gets P_{C,k}$
  \State $\exists \quad x_k^{c1} \in  I_{c1} \quad (c1 \in [1,D])$ // find the group contain $x_k^{c1}$
  \State $\exists \quad x_k^{c2} \in  I_{c2} \quad (c2 \in [1,D])$
  \State $I_{c1} \gets I_{c1} \cup I_{c2}$
  \State $I_{c2} \gets \varnothing$
  \EndFor
  \State $k\gets 1$
  \While{$k \leq [D/3]$}
  \State $[x_k^{s1},x_k^{s2}]\gets P_{S,k}$
  \State $\exists \quad x_k^{s1} \in  I_{s1} \quad (s1 \in [1,D])$
  \State $\exists \quad x_k^{s2} \in  I_{s2} \quad (s2 \in [1,D])$
  \If{ $||I_{s2} \cup I_{s2}|| \leq \Delta \text{ and } [x_k^{s1},x_k^{s2}] \notin P_{C}$} 
    \State $I_{s1} \gets I_{s2} \cup I_{s2}$
    \State $I_{s2} \gets \varnothing$
    \State $k\gets k+1$
  \EndIf
  \EndWhile
  \State \Return {${\bf{I}}$}
  \end{algorithmic}  
  \end{algorithm}  
\subsubsection{Searching scope aggregation}
\par 
In the DA-EGO algorithm, each variable is assigned an independent search range. 
However, when multiple variables are allocated to the same optimization task, the interaction relationships need to be considered to adjust the search range of the subtasks. 
Based on the knowledge of design space, the search range of sub-problems can be reduced, which can significantly enhance optimization efficiency and knowledge mining accuracy. 
The adjustment of the search range is based on accumulated experience. 
If similar sub-tasks have been previously solved (meaning sub-problems with the same variables, but different elite points), the new search can be based on previous experience. 
Conversely, if a new sub-task is unknown (meaning new variables are added to the group), the new problem must be thoroughly explored. Equation (\ref{range}) shows the specific method.
\begin{equation}\label{range}
\begin{small}
{S_j} = \left\{ 
\begin{array}{*{20}{c}}
  {[l_{k},u_{k}]}^{\|{I_j}\|} \quad  (x_{k} \in {I_j})   \quad
(\text{if the variable in}\; {I_j} \;\text{remain the same}) \\
  {[0,1]}^{\|{I_j}\|}
\qquad \qquad \qquad \qquad \qquad (\text{if the variable in}\; {I_j} \;\rm{changed})
\end{array}
\right.
\end{small}
\end{equation}
Where $l_k$ and $u_k$ are the lower and upper boundary values of the $k^{th}$ variable in the $j^{th}$ subset design space.
These variable ranges are obtained in the previous cycle of sub-tasks and recorded in the boundary table as shown in Fig.\ref{dataflow}. 
\subsubsection{Generate new sub-tasks}
When the elite-point, decomposition scheme, and search scope are all updated with the knowledge aggregate method, new sub-tasks are created based on this new knowledge of the whole design space. 
During the optimization of the samples in the subspace, only the values of the variables corresponding to the subspace change, while the remaining variables retain the values of the elite-point. 
Following these steps, sub-problem construction for the new optimization cycle is complete. 
A thorough comprehension of the original problem leads to strong interactions within the sub-problems but weak interactions between them, resulting in a search boundary effectively narrowed down to the optimal region. 
This procedure fully illustrates how design space knowledge can be used to effectively construct sub-optimization problems.
\subsection{Sub-task optimization and knowledge mining} 
\par
The sub-task optimization and knowledge mining module is faced with an independent sub-task as shown in Equation (\ref{sub}), which only contains several variables of the original problem, so its dimension is low. 
\begin{equation}\label{sub}
\begin{array}{l}
\min f_j^{sub}({\bf{x}}) = \min f(G({\bf{x}},{I_j},{x^*}))\\
{\rm{s}}{\rm{.t}}\qquad {\rm{.     }}{l_k} \le {{\bf{x}}_k} \le {u_k} \quad (x_{k} \in {I_j})
\end{array}
\end{equation}
Where the function $G(\cdot)$ denotes the mapping relationship between subspace and global space, which has been defined in Equation \ref{gen_sub}.
The design space of sub-problems is totally determined by the elite-point $x^*$, decomposition scheme(which is expressed as group index $\bf{I}$), and search scope $[L, U]$ together. 
\subsubsection{Sub-task optimization}
\par
Firstly, the EGO algorithm is used to search for the optimal sample in the sub-problem space. 
As described in Section 2.3, the EGO algorithm usually uses the maximum expected improvement (EI) criterion to achieve a balance between local search and global exploration. 
Therefore, the accuracy of the Kriging surrogate established will be continuously improved with new samples added. 
In this paper, the initial sampling number of the EGO algorithm is set as $3d$, the maximum number of iterations is set as $6d$, and here $d$ is the subspace dimension. 
After the optimization, the optimal sample and the predictive surrogate $\hat f_{sub}^{KG}$ of the subspace can be obtained.
\subsubsection{Sub-task knowledge mining}
\par
After obtaining the optimal sample of the subproblem, the search scope and variable interaction of the subproblem need to be obtained through knowledge mining technology.
The search boundary of the sub-problem can be redefined based on the obtained surrogate $\hat f_{sub}^{KG}$. 
In the DA-EGO algorithm, the space reduction method is used to learn the new boundary in the sub-problem. 
The space reduction method can make the optimization search focus on the potential optimal region and avoid unnecessary calculation costs in the non-important region. 
The detailed process of the space reduction is expressed as pseudocode in Algo. 2.
And in this paper, the sampling number $N_{SR}$ in surrogate $\hat f_{sub}^{KG}$ is set as 10000, and the space reducing ratio $\omega$ is set as 0.5.
\begin{algorithm}  
\caption{Space reduction in the $j^{\text{th}}$ sub-task}  
\begin{algorithmic}[1]
\Require 
\Statex the sampling number($N_{SR}$);
the surrogate $\hat f_{sub}^{KG}$;
the current searching range of $j^{\text{th}}$ sub-task (${[l_{k},u_{k}]\quad(x_k \in {I_j})}$); 
the space reducing ratio ($\omega$);
\Ensure 
\Statex  the new searching range of $j^{\text{th}}$ sub-task (${[l_{k},u_{k}]\quad(x_k \in {I_j})}$); 
\State $[X^{1},\cdots,X^{N_{SR}}] \gets$ generate $N_{SR}$ samples in the range ${[l_{k},u_{k}],\quad(x_k \in {I_j})}$
\For{$k= 1 \to N$} 
\State${\hat{y}^k \gets \hat f_{sub}^{KG}(X^{k})}$
\EndFor
\State $[X^{\rm{rank}1},\cdots,X^{\rm{rank}N_{SR}}] \gets$ sort $[X^{1},\cdots,X^{N_{SR}}]$ in ascending order according to $\hat{y}$
\State${M \gets [\omega \cdot N_{SR}]}$
\For{$k= 1 \to D$} 
\State${{l_k} \gets \min (x_k^{{\rm{rank}}1}, x_k^{{\rm{rank}}2}, \cdots, x_k^{{\rm{rank}}M})}$
\State${{u_k} \gets \max (x_k^{{\rm{rank}}1}, x_k^{{\rm{rank}}2}, \cdots, x_k^{{\rm{rank}}M})}$
\EndFor
\State \Return{$[l_{k},u_{k}]\quad(x_k \in {I_j})$} 
\end{algorithmic}  
\end{algorithm}  
\par 
In the DA-EGO algorithm, a common interaction analysis method, the perturbation method is used to analyze the interaction between variables in the same sub-tasks.
The perturbation method evaluates variable relationships one pair at a time. 
Consequently, a problem with $d$ variables requires the evaluation of $d(d-1)/2$ variable pairs separately. 
Moreover, the perturbation analysis methodology, based on a surrogate, can improve the accuracy of results by removing regions of the surrogate model with low accuracy. 
According to literature~\cite{kangEfficientHighdimensionalMetamodeling2020}, limiting the use of the perturbation method to the ${[0.1,0.9]^D}$ range significantly enhances its precision. Thus, restricting the perturbation analysis sampling within a reasonable range can improve its accuracy.
According to the established subspace kriging model $\hat f_{sub}^{KG}$, the perturbation method is used to confirm whether the selected variable pairs have interactive relations. Equation (\ref{perturbation}) is the specific method of perturbation analysis, in which parameter values are consistent with those in the literature ~\cite{kangEfficientHighdimensionalMetamodeling2020}, set $N_{PA}=10000$, and $h=0.0001$. 
\begin{equation}\label{perturbation}
	\begin{array}{l}
	{H_{ij}} = \frac{1}{{{N_{{\rm{PA}}}}{h^2}(\max (\hat f_{sub}^{KG}) - \min (\hat f_{sub}^{KG}))}}\\
	\sum\limits_{k = 1}^{{N_{{\rm{PA}}}}} \mid  \hat f_{sub}^{KG}\left( {{x^{k}_{1}},{x^{k}_{2}}, \ldots ,{x^{k}_{i}} + h, \ldots ,{x^{k}_{j}} + h, \ldots ,{x^{k}_{d}}} \right)\\
	- \hat f_{sub}^{KG}\left( {{x^{k}_{1}},{x^{k}_{2}}, \ldots ,{x^{k}_{i}} + h, \ldots ,{x^{k}_{j}}, \ldots ,{x^{k}_{d}}} \right)\\
	- \hat f_{sub}^{KG}\left( {{x^{k}_{1}},{x^{k}_{2}}, \ldots ,{x^{k}_{i}}, \ldots ,{x^{k}_{j}} + h, \ldots ,{x^{k}_{d}}} \right)\\
	+ \hat f_{sub}^{KG}\left( {{x^{k}_{1}},{x^{k}_{2}}, \ldots ,{x^{k}_{i}}, \ldots ,{x^{k}_{j}}, \ldots ,{x^{k}_{d}}} \right)\mid 
	\end{array}
\end{equation}
If the condition $H_{ij} \geq 0.2$ is met, the interaction between variable $x_i$ and $x_j$ is confirmed. 
Then, these confirmed interacted variable pairs are recorded and feedback to the problem decomposition part.
\par
With the completion of the above steps, all evaluation samples of the sub-problem, new search boundaries, and interacted pairs of variables can be obtained. 
As shown in Fig.\ref{dataflow}, this knowledge will be extracted from each sub-tasks and concentrated into the three tables storing the original problem information: 
(\romannumeral1) the sample points of the subspace will be added to the evaluated data table, 
(\romannumeral2) the variable pairs that actually have interactive relations will be added to the interactive table, 
(\romannumeral3) the obtained search boundary will replace the original boundary in the search boundary table. After obtaining these feedbacks, the knowledge of the global problem becomes more accurate and rich, which is also conducive to the construction of the following sub-problems.
\par
As knowledge is received in each cycle, problems are decomposed and sub-problems optimized repeatedly while gaining increasingly better samples, interaction structures of original high-dimensional problems become clearer. 
This gradual exploration also enables smaller search scopes, which are ideal for enhancing local search capabilities.
\section{Numerical Experimental Study}
In this section, the performance of our proposed DA-EGO algorithm is evaluated on 21 benchmark instances. 
And the DA-EGO is also compared with the state-of-the-art algorithms.
\subsection{Experimental settings}
\par 
To show the effectiveness of the proposed algorithm, it is compared against the following three kinds of algorithms:
\par
(1)
The first kind is the model-free algorithms including the classical genetic algorithm (GA)~\citep{mitchellIntroductionGeneticAlgorithms1998}, differential evolution algorithm (DE)~\citep{stornDifferentialEvolutionSimple1997} and a recent proposed gaining-sharing knowledge based algorithm (GSK)~\citep{mohamedGainingsharingKnowledgeBased2020}. 
For these 3 algorithms, their population size is set to be 50 and the remaining variables are in consistent with the default values shown in the related papers. 
\par
(2)
The second kind is the surrogate assisted genetic algorithms including the incremental kriging-assisted evolutionary algorithm (IKAEA)~\citep{zhanFastKrigingAssistedEvolutionary2021}
and generalized surrogate-assisted genetic algorithm (GSGA)~\citep{caiEfficientSurrogateassistedParticle2019}.
Note that the IKAEA and GSGA are recently proposed SAEAs for high-dimensional optimization problems.
We follow the default settings in the papers of IKAEA and GSGA for the following tests.
\par
(3)
The third kind is the surrogate-based algorithm, Nash-EGO, which has mentioned in introduction.
Besides, the RG-EGO, a variant of the DA-EGO algorithm is also joined in the comparison.
The RG-EGO used random grouping decomposition strategy while keep all other setting as same as the DA-EGO.
Therefore, the comparing the performance of the two can clearly reflect the effect of the novel interaction analysis strategy proposed in the paper.
\par
One of the principle of comparison algorithm selection is the consistency of the applicable dimensions and the range of sample sizes.
So, some algorithms with similar mechanism but different scope of application (i.e. SEE, OMID, CCVIL) are not involved in the comparison.
\par 
In this paper, the tests for all the benchmark problems are repeated 20 times.
In the $i^{\text{th}}$ test of each benchmark function $(i=1 \cdots 20)$, the initial sample distribution (or called initial population) of all algorithms is the same.
The termination condition is the number of function evaluations (NFE) reaches to 1500.
\subsection{Benchmark functions}
\par
All these benchmark functions used in this paper are selected from the CEC's 2010 special session~\citep{tangBenchmarkFunctionsCEC2009}. 
The characteristics of these benchmark functions are summarized in Table \ref{benchmarkFeature}.
And their formulations are shown in Table \ref{benchmarkEquation}.
\begin{table}[htbp]
  \centering
  \caption{Benchmark functions}
    \begin{tabular}{lllll}
    \toprule
   Index & Dim & Name & Property & Range\\
    \midrule
    F1    & 30/60/90    & {Shifted Elliptic} & separable   &$[-10,10]^{D}$  \\
    F2    & 30/60/90    & {Shifted Rastrigin}  & separable        & $[-5,5]^{D}$ \\
    F3    & 30/60/90    & {Shifted Ackley}   & separable       &$[-32,32]^{D}$ \\
    F4    & 30/60/90    & {Rotated Elliptic}    & partially-separable    &$[-10,10]^{D}$  \\
    F5    & 30/60/90    & {Rotated Rastrigin}  & partially-separable     & $[-5,5]^{D}$  \\
    F6    & 30/60/90    & {Rotated Ackley}  & partially-separable      &$[-32,32]^{D}$  \\
    F7    & 30/60/90    & {Shifted Rosenbrock} & non-separable & $[-10,10]^{D}$ \\
    \bottomrule
    \end{tabular}%
    \label{benchmarkFeature}
\end{table}%

\begin{table}[htbp]
  \centering
  \caption{The equations of benchmark function}
  \resizebox{\linewidth}{!}{
    \begin{tabular}{ll}
    \toprule
     Index & \multicolumn{1}{l}{Description} \\
    \midrule
    F1    &  $f(x) = F_{\text {elliptic }}(Z_{sft}(x))$\\
    F2    &  $f(x) = F_{\text {rastrigin }}(Z_{sft}(x))$\\
    F3    &  $f(x) = F_{\text {ackley }}(Z_{sft}(x))$\\
    F4    &  $f(x) = F_{\text {elliptic }}(Z_{rot}(x))$\\
    F5    &  $f(x) = F_{\text {rastrigin }}(Z_{rot}(x))$\\
    F6    &  $f(x) = F_{\text {ackley }}(Z_{rot}(x))$\\
    F7    & $f(x)=F_{\text{Rosenbrock}}(Z_{sft}(x))$\\
\midrule
    $F_{\text {elliptic }}$ & $ f(z)=\sum_{i=1}^{D}\left(10^{6}\right)^{\frac{i-1}{D-1}} z_{i}^{2}$ \\
    $F_{\text {rastrigin }}$  & $f(z) = \sum_{i=1}^{D}\left[z_{i}^{2}-10 \cos \left(2 \pi z_{i}\right)+10\right]$ \\
    $F_{\text {ackley }}$  &  
\tabincell{c}{$f(z) =-20 \exp \left(-0.2 \sqrt{\frac{1}{D} \sum_{i=1}^{D} z_{i}^{2}}\right)$
$-\exp \left(\frac{1}{D} \sum_{i=1}^{D} \cos \left(2 \pi z_{i}\right)\right)+20+e $}\\
    $F_{\text{Rosenbrock}}$ & $f(z)=\sum_{i=1}^{D-1}\left[100\left(z_i^2-z_{i+1}\right)^2+\left(z_i-1\right)^2\right]$\\
    $Z_{sft}$   &  $f(x) = (x-O)[P_{1}:P_{D}]$\\
    $Z_{rot}$   &  \footnotesize{$\begin{cases} f(x)[1+5n:5+5n] = Z_{sft}(x)[1+5n:5+5n]*M &  (\text{if } n =0,\cdots,d/10-1)  \\ 
    f(x)[1+5n:5+5n] = Z_{sft}(x)[1+5n:5+5n] &(\text{if } n =d/10,\cdots,d/5) \end{cases}$} \\
    \bottomrule
    \end{tabular}}
  \label{benchmarkEquation}%
\end{table}%
\par
Note that the variable interactions has a great impact on the performance of the optimization algorithm. 
The test suite contains nine separable instances (F1--F3), nine partially separable instances (F4--F6), and three non-separable shifted Rosenbrock instances (F7), giving 21 instances in total. Each of the seven functions is tested at 30, 60, and 90 dimensions.
Specifically, the separable functions are the functions whose design variables are independent.
In contrast, the design variables of the non-separable functions are highly interacted.
More specifically, we follow the operations in~\citep{tangBenchmarkFunctionsCEC2009}, such as shifting, rearrangement and rotation to build the benchmark functions, as shown in Table \ref{benchmarkEquation}.
In particularly, the expressions of $O$, $M$, and $P$ can be found in our previous work ~\cite{wangKTEGOKnowledgeTransfer2022a}.
\subsection{Result of benchmark functions}
\subsubsection{Comparison result of 30-D benchmark functions}
\par
\begin{figure}[htp]
\centering

\subfigure[F1]{
\begin{minipage}[t]{0.33\linewidth}
\centering
\includegraphics[width=1\textwidth]{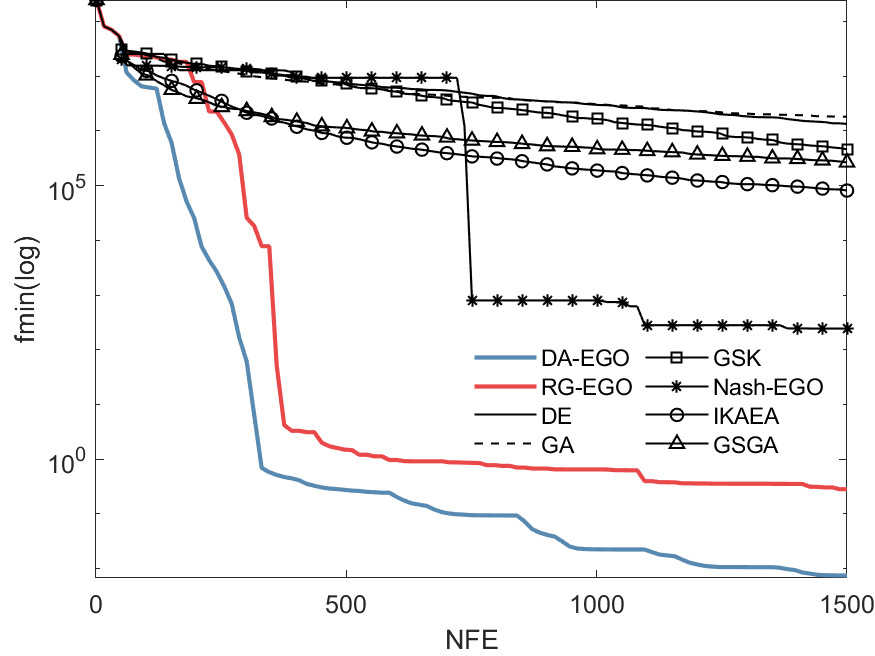}
\end{minipage}%
}%
\subfigure[F2]{
\begin{minipage}[t]{0.33\linewidth}
\centering
\includegraphics[width=1\textwidth]{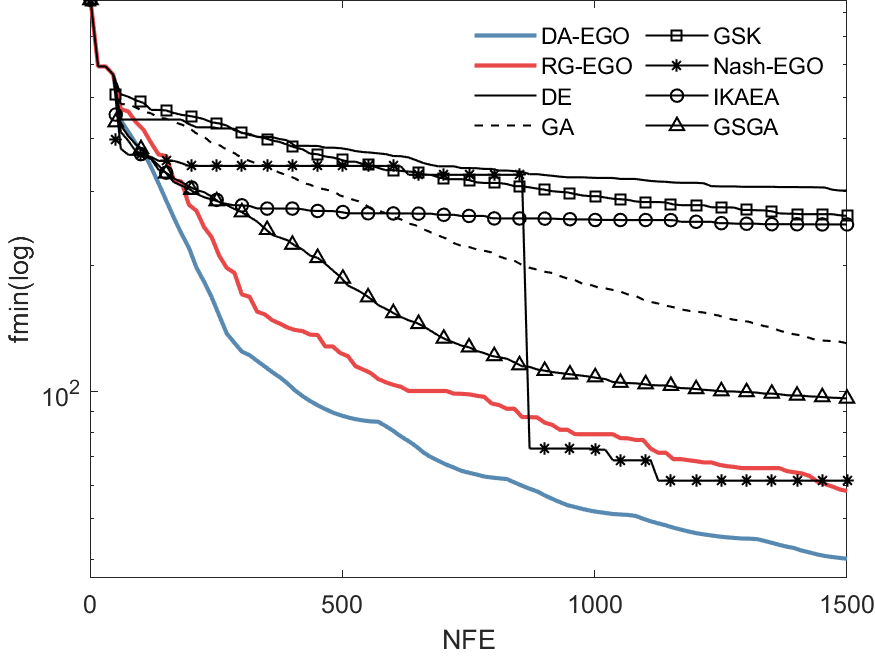}
\end{minipage}
}%
\subfigure[F3]{
\begin{minipage}[t]{0.33\linewidth}
\centering
\includegraphics[width=1\textwidth]{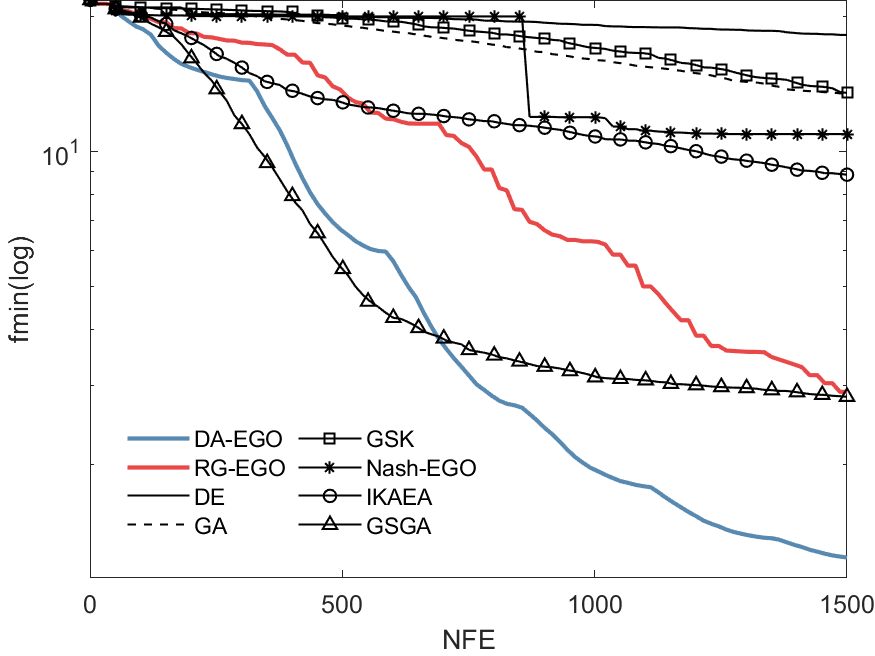}
\end{minipage}
}%

\subfigure[F4]{
\begin{minipage}[t]{0.33\linewidth}
\centering
\includegraphics[width=1\textwidth]{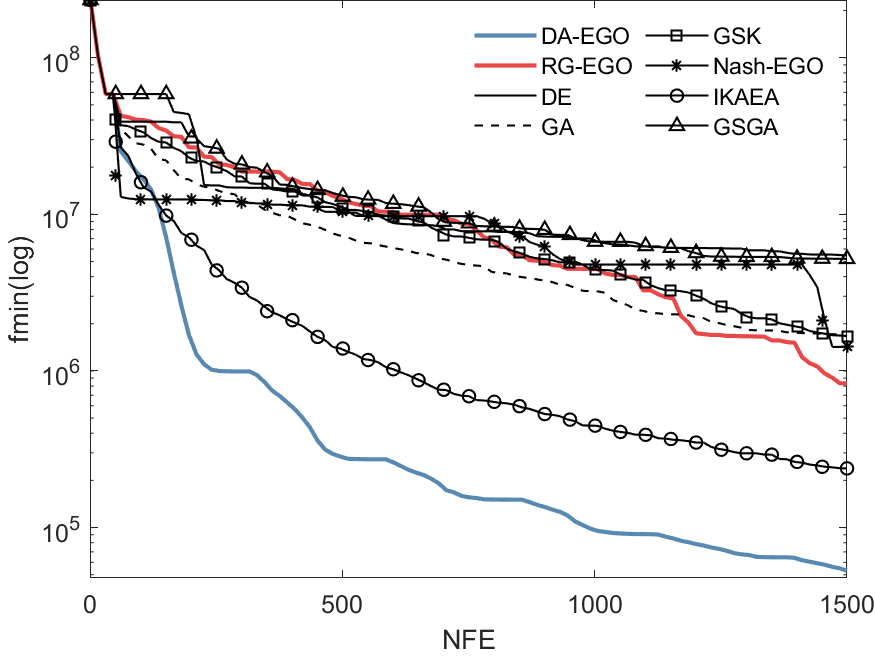}
\end{minipage}%
}%
\subfigure[F5]{
\begin{minipage}[t]{0.33\linewidth}
\centering
\includegraphics[width=1\textwidth]{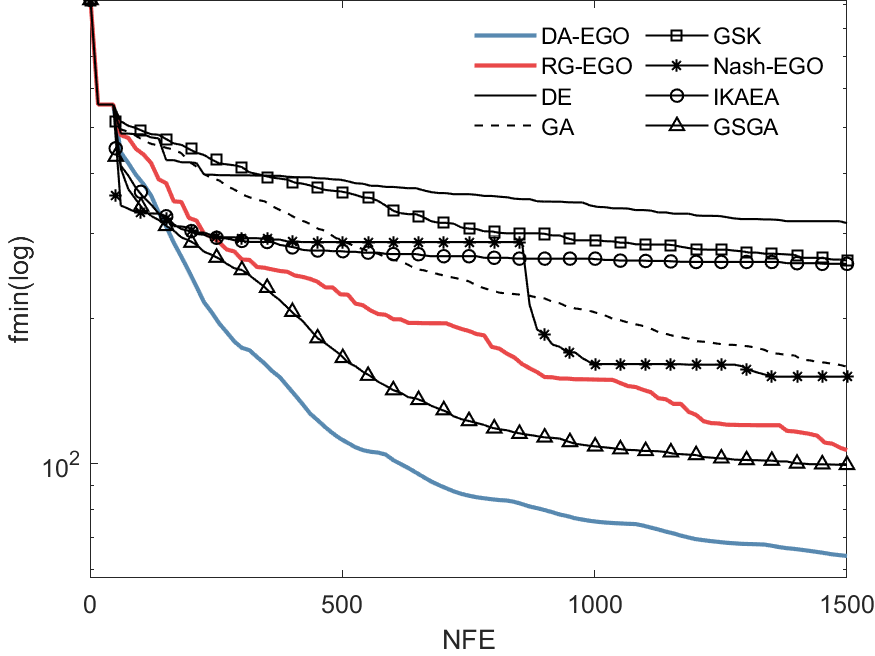}
\end{minipage}
}%
\subfigure[F6]{
\begin{minipage}[t]{0.33\linewidth}
\centering
\includegraphics[width=1\textwidth]{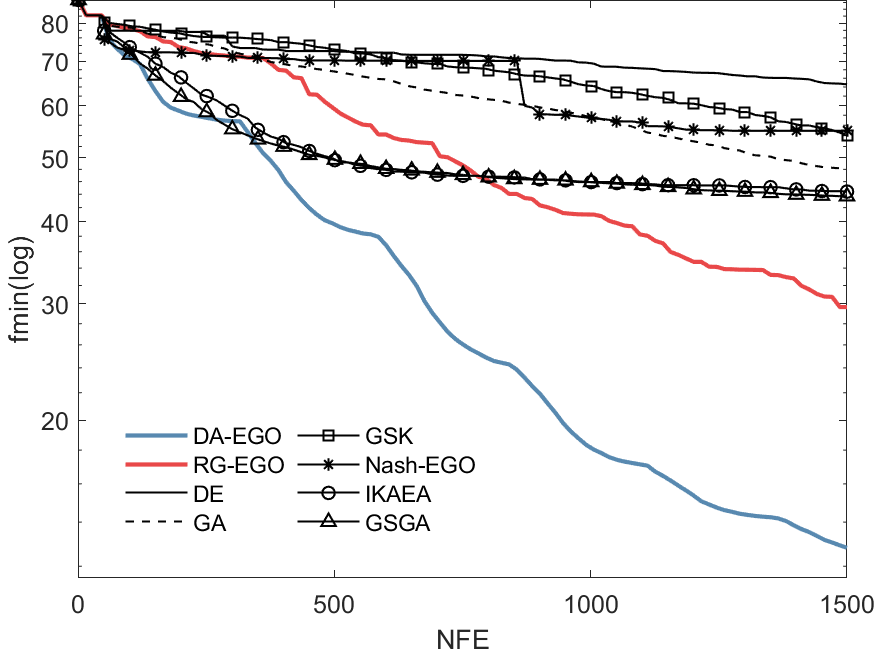}
\end{minipage}
}%
\centering
\caption{Average convergence curves on F1--F6 at 30 dimensions.}
\label{30Dfigure}
\end{figure}

 
The convergence curves for F1--F6 at 30 dimensions are shown in Fig.~\ref{30Dfigure}; the corresponding F7 curve is included in Fig.~\ref{F7figure}. 
The details of the optimization results are listed in Table \ref{30Dtable},
where ``Std'' represents standard deviation, and the ``Wilcoxon'' represent Wilcoxon signed rank test ~\citep{derracPracticalTutorialUse2011}, the symbols `+', `$-$', and `$\approx$' indicate the DA-EGO algorithm is significantly better than, significantly worse than, or comparable to the compared algorithm.
And the optimal mean result of each function is marked in bold.
\par
Table~\ref{30Dtable} shows that DA-EGO performs significantly better than the seven comparison algorithms on F1--F6 at 30 dimensions. For F7, its Wilcoxon comparison is statistically comparable to IKAEA and GSGA, while it is significantly better than the other comparison algorithms. GSGA has a slightly lower reported mean on F7. 
The values of standard deviation reflect the robustness of the algorithm for the initial distribution. %
The reported standard deviations vary across algorithms and functions; the F7 results do not support a universal robustness advantage for DA-EGO.
\par
Here, benchmark functions are discussed separately according to the separability.
In Fig.\ref{30Dfigure}(a), (b), and (c), benchmark functions F1, F2, and F3 are separable functions. 
Those decomposition based algorithms such as Nash-EGO, DA-EGO, and the RG-EGO have advantages as the combination of all the sub-optimization results is bound to get a better solution in separable functions.
However, the GSGA algorithm also has great performance in multi-modal functions F2 and F3.
Compare the DA-EGO and the RG-EGO in detail, it is found that both of them have similar convergence curves, but DA-EGO has higher efficiency due to the existence of space reduction strategy.
\par 
On the other hand, in Fig.\ref{30Dfigure}(d), (e), and (f), benchmark functions F4, F5, and F6 are partially-separable functions, half of their variables are strongly interacted with each other.
For such problems, the key to the efficient search of the decomposition based algorithm is whether the problem can be decomposed accurately according to the interaction.
Two algorithms using manually decomposition strategy, Nash-EGO and RG-EGO, lose their advantages 
compared with other algorithms. 
In contrast, the convergence rate and the final solutions of DA-EGO are still significantly better than the compared algorithms.
\begingroup
\scriptsize
\setlength{\tabcolsep}{3.3pt}
\renewcommand{\arraystretch}{1.06}
\begin{longtable}{llrrrrc}
\caption{Optimization results of 8 algorithms on the seven 30-D benchmark functions.}\label{30Dtable}\\
\toprule
Func. & Algorithm & Best & Worst & Mean & Std & Wilcoxon \\
\midrule
\endfirsthead
\multicolumn{7}{c}{Table~\ref{30Dtable} continued}\\
\toprule
Func. & Algorithm & Best & Worst & Mean & Std & Wilcoxon \\
\midrule
\endhead
\midrule
\multicolumn{7}{r}{Continued on next page}\\
\endfoot
\bottomrule
\endlastfoot
F1 & DA-EGO & 2.226E-03 & 4.311E-02 & \textbf{1.730E-02} & 1.263E-02 & N/A \\
 & RG-EGO & 1.170E-01 & 6.607E-01 & 3.295E-01 & 1.591E-01 & $+$ \\
 & DE & 7.722E+05 & 2.090E+06 & 1.336E+06 & 3.541E+05 & $+$ \\
 & GA & 2.339E+05 & 5.043E+06 & 1.814E+06 & 1.289E+06 & $+$ \\
 & GSK & 1.758E+05 & 1.470E+06 & 4.590E+05 & 2.633E+05 & $+$ \\
 & Nash-EGO & 1.359E+02 & 3.958E+02 & 2.442E+02 & 1.032E+02 & $+$ \\
 & IKAEA & 3.704E+04 & 2.456E+05 & 8.105E+04 & 4.950E+04 & $+$ \\
 & GSGA & 1.392E+05 & 5.068E+05 & 2.641E+05 & 1.005E+05 & $+$ \\
\midrule
F2 & DA-EGO & 2.685E+01 & 5.859E+01 & \textbf{4.653E+01} & 8.412E+00 & N/A \\
 & RG-EGO & 4.730E+01 & 8.265E+01 & 6.328E+01 & 1.007E+01 & $+$ \\
 & DE & 2.651E+02 & 3.493E+02 & 3.010E+02 & 2.188E+01 & $+$ \\
 & GA & 8.565E+01 & 1.730E+02 & 1.304E+02 & 2.604E+01 & $+$ \\
 & GSK & 2.226E+02 & 2.991E+02 & 2.622E+02 & 2.271E+01 & $+$ \\
 & Nash-EGO & 6.146E+01 & 6.146E+01 & 6.146E+01 & 7.290E-15 & $+$ \\
 & IKAEA & 2.058E+02 & 2.763E+02 & 2.501E+02 & 2.048E+01 & $+$ \\
 & GSGA & 6.408E+01 & 1.458E+02 & 9.642E+01 & 2.111E+01 & $+$ \\
\midrule
F3 & DA-EGO & 1.234E+00 & 2.126E+00 & \textbf{1.522E+00} & 1.967E-01 & N/A \\
 & RG-EGO & 1.896E+00 & 4.016E+00 & 3.152E+00 & 6.509E-01 & $+$ \\
 & DE & 1.659E+01 & 1.894E+01 & 1.816E+01 & 5.639E-01 & $+$ \\
 & GA & 1.013E+01 & 1.605E+01 & 1.341E+01 & 2.257E+00 & $+$ \\
 & GSK & 1.157E+01 & 1.542E+01 & 1.348E+01 & 1.023E+00 & $+$ \\
 & Nash-EGO & 1.034E+01 & 1.106E+01 & 1.092E+01 & 2.953E-01 & $+$ \\
 & IKAEA & 7.189E-01 & 1.904E+01 & 8.861E+00 & 4.752E+00 & $+$ \\
 & GSGA & 7.281E-01 & 4.747E+00 & 2.834E+00 & 1.197E+00 & $+$ \\
\midrule
F4 & DA-EGO & 4.114E+03 & 2.843E+05 & \textbf{7.672E+04} & 8.002E+04 & N/A \\
 & RG-EGO & 9.764E+04 & 3.615E+06 & 1.285E+06 & 1.302E+06 & $+$ \\
 & DE & 3.557E+06 & 7.844E+06 & 5.463E+06 & 1.130E+06 & $+$ \\
 & GA & 3.842E+05 & 3.715E+06 & 1.670E+06 & 9.448E+05 & $+$ \\
 & GSK & 9.913E+05 & 2.828E+06 & 1.667E+06 & 4.954E+05 & $+$ \\
 & Nash-EGO & 1.394E+06 & 1.450E+06 & 1.422E+06 & 2.197E+04 & $+$ \\
 & IKAEA & 1.052E+05 & 5.353E+05 & 2.387E+05 & 1.053E+05 & $+$ \\
 & GSGA & 2.373E+05 & 9.421E+07 & 5.194E+06 & 2.095E+07 & $+$ \\
\midrule
F5 & DA-EGO & 4.203E+01 & 9.876E+01 & \textbf{6.992E+01} & 1.535E+01 & N/A \\
 & RG-EGO & 8.555E+01 & 1.495E+02 & 1.135E+02 & 2.291E+01 & $+$ \\
 & DE & 2.707E+02 & 3.736E+02 & 3.155E+02 & 2.574E+01 & $+$ \\
 & GA & 1.167E+02 & 1.950E+02 & 1.586E+02 & 2.357E+01 & $+$ \\
 & GSK & 2.246E+02 & 2.912E+02 & 2.638E+02 & 1.660E+01 & $+$ \\
 & Nash-EGO & 1.282E+02 & 1.649E+02 & 1.512E+02 & 1.319E+01 & $+$ \\
 & IKAEA & 2.340E+02 & 3.062E+02 & 2.596E+02 & 1.659E+01 & $+$ \\
 & GSGA & 5.939E+01 & 1.320E+02 & 9.909E+01 & 1.874E+01 & $+$ \\
\midrule
F6 & DA-EGO & 7.191E+00 & 4.419E+01 & \textbf{1.534E+01} & 9.795E+00 & N/A \\
 & RG-EGO & 1.179E+01 & 4.094E+01 & 3.094E+01 & 8.138E+00 & $+$ \\
 & DE & 5.568E+01 & 7.117E+01 & 6.460E+01 & 3.402E+00 & $+$ \\
 & GA & 3.629E+01 & 6.150E+01 & 4.813E+01 & 8.936E+00 & $+$ \\
 & GSK & 4.550E+01 & 6.118E+01 & 5.400E+01 & 3.952E+00 & $+$ \\
 & Nash-EGO & 5.179E+01 & 5.996E+01 & 5.493E+01 & 3.127E+00 & $+$ \\
 & IKAEA & 2.415E+01 & 6.870E+01 & 4.442E+01 & 1.026E+01 & $+$ \\
 & GSGA & 2.700E+01 & 6.235E+01 & 4.370E+01 & 1.087E+01 & $+$ \\
\midrule
F7 & DA-EGO & 1.906E+02 & 3.646E+03 & 3.178E+02 & 1.963E+03 & N/A \\
 & RG-EGO & 5.451E+03 & 1.623E+04 & 1.583E+03 & 3.076E+03 & $+$ \\
 & DE & 2.394E+05 & 9.609E+05 & 5.187E+05 & 1.930E+05 & $+$ \\
 & GA & 7.232E+03 & 2.942E+04 & 1.444E+04 & 8.182E+03 & $+$ \\
 & GSK & 1.217E+04 & 1.136E+05 & 4.437E+04 & 2.838E+04 & $+$ \\
 & Nash-EGO & 5.506E+03 & 1.944E+04 & 6.241E+03 & 3.819E+03 & $+$ \\
 & IKAEA & 6.009E+02 & 2.276E+03 & 1.107E+03 & 5.688E+02 & $\approx$ \\
 & GSGA & 1.371E+02 & 2.615E+03 & \textbf{3.160E+02} & 1.178E+03 & $\approx$ \\
\end{longtable}
\endgroup

\subsubsection{Comparison result of 60-D benchmark functions}
\par
\begin{figure}[htp]
  \centering
  
  \subfigure[F1]{
  \begin{minipage}[t]{0.33\linewidth}
  \centering
  \includegraphics[width=1\textwidth]{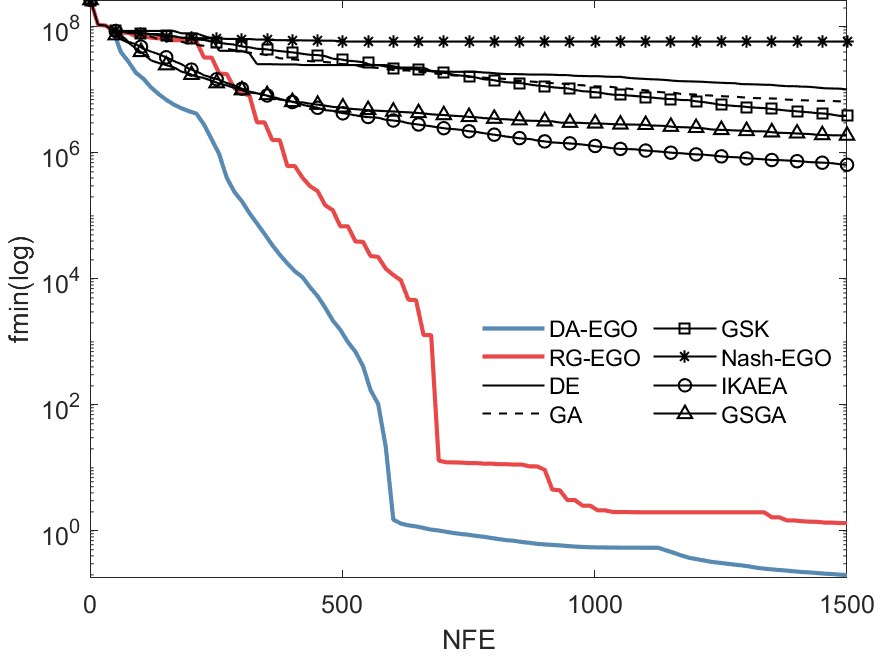}
  \end{minipage}%
  }%
  \subfigure[F2]{
  \begin{minipage}[t]{0.33\linewidth}
  \centering
  \includegraphics[width=1\textwidth]{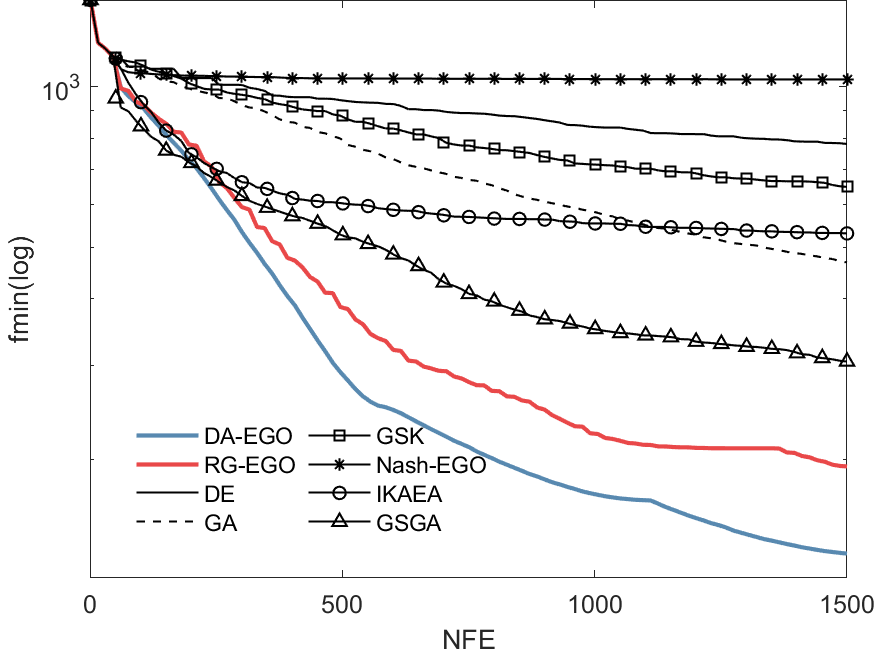}
  \end{minipage}
  }%
  \subfigure[F3]{
  \begin{minipage}[t]{0.33\linewidth}
  \centering
  \includegraphics[width=1\textwidth]{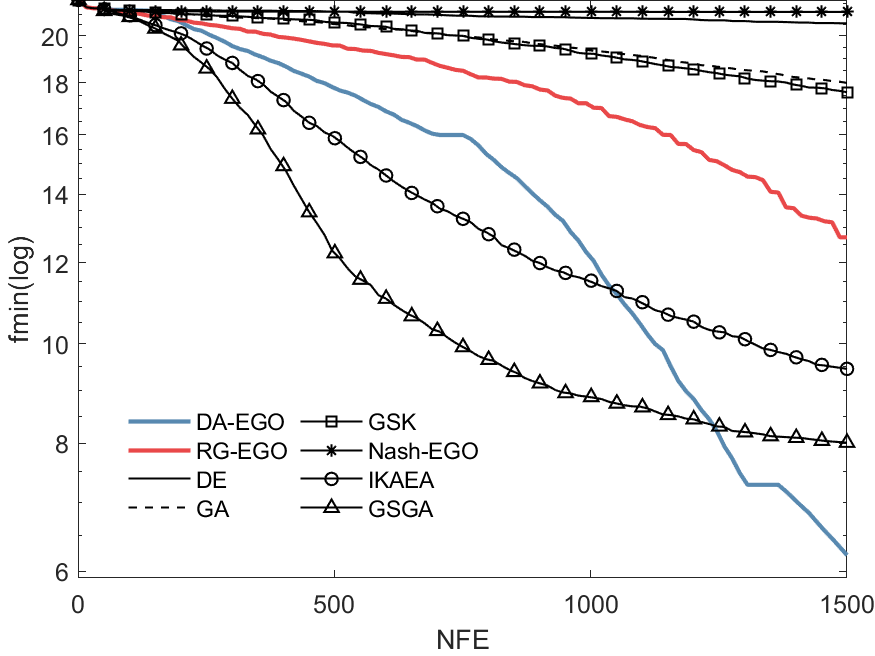}
  \end{minipage}
  }%

  \subfigure[F4]{
  \begin{minipage}[t]{0.33\linewidth}
  \centering
  \includegraphics[width=1\textwidth]{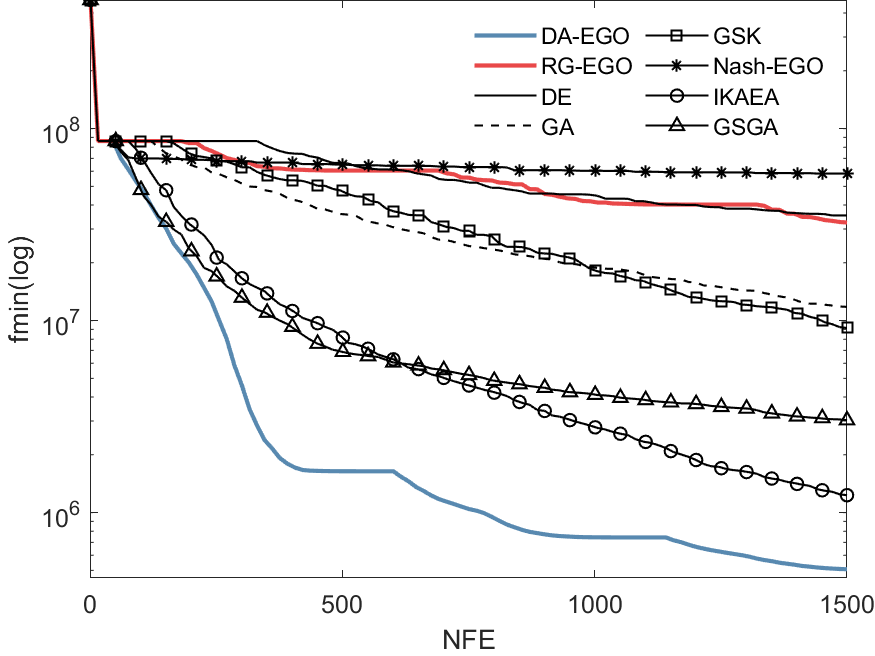}
  \end{minipage}%
  }%
  \subfigure[F5]{
  \begin{minipage}[t]{0.33\linewidth}
  \centering
  \includegraphics[width=1\textwidth]{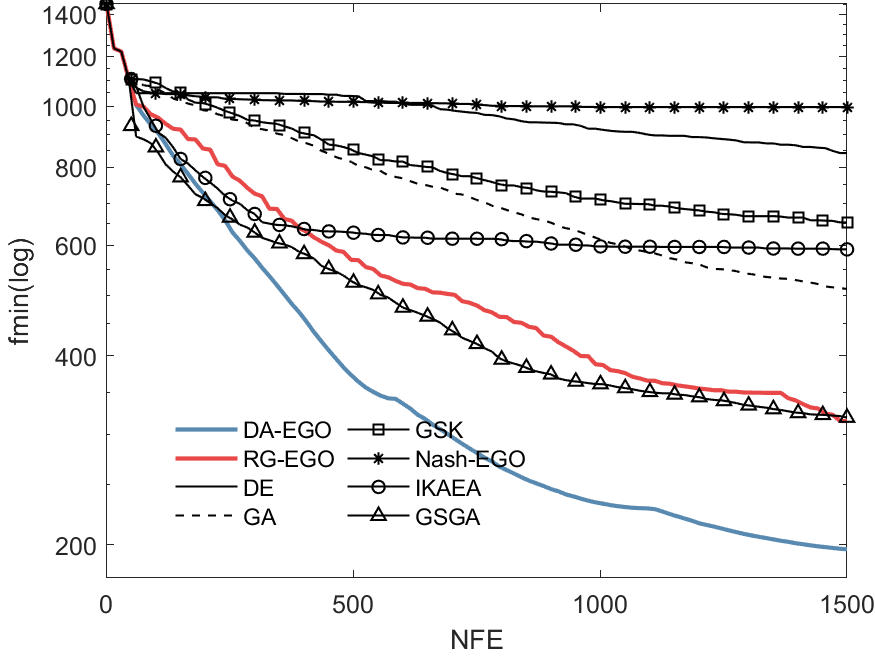}
  \end{minipage}
  }%
  \subfigure[F6]{
  \begin{minipage}[t]{0.33\linewidth}
  \centering
  \includegraphics[width=1\textwidth]{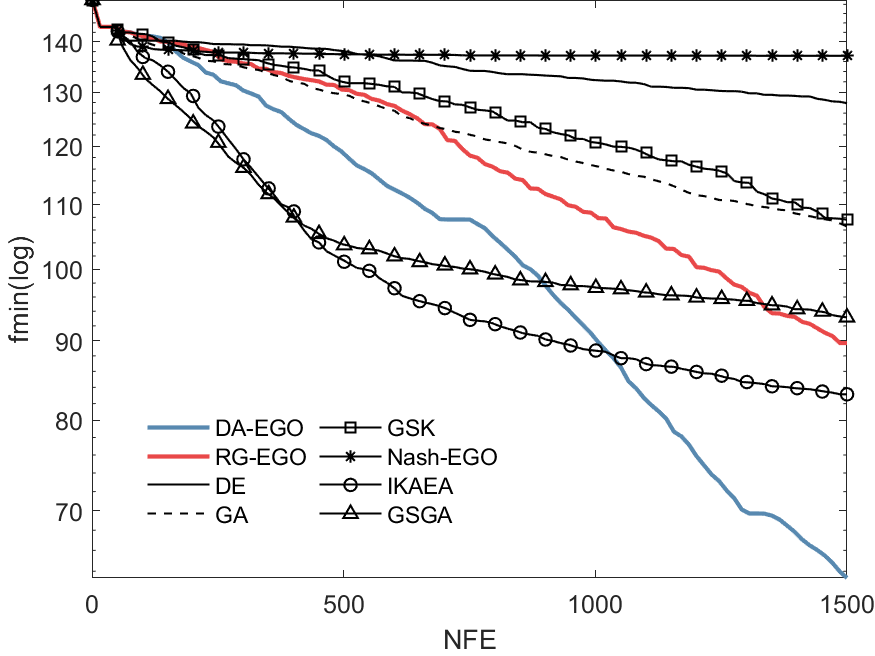}
  \end{minipage}
  }%
  
  \centering
  \caption{Average convergence curves on F1--F6 at 60 dimensions.}
  \label{60Dfigure}
  \end{figure}

 
The convergence curves for F1--F6 at 60 dimensions are shown in Fig.~\ref{60Dfigure}, with F7 shown in Fig.~\ref{F7figure}.
The details of the optimization results are listed in Table \ref{60Dtable}.
\par
Table~\ref{60Dtable} compares seven algorithms at 60 dimensions. DA-EGO has the lowest reported mean on F1--F6, although the F3 comparison with GSGA is statistically comparable. On F7, GSGA has a lower reported mean (9.931E+03 versus 1.539E+04) and the Wilcoxon result favors GSGA. 
Similar with the result in the test of 30-dimensional cases, both DA-EGO and RG-EGO have clear advantages in the test on F1 function.
However, in the test on F4 function, a variant of F1 with strong interactions, the DA-EGO still has high efficiency but the RG-EGO is difficult to converge.
The above huge different between two cases shows the role of proposed screening and identification strategy in the face of strong interaction problems.
The DA-EGO algorithm accurately decomposes the original problems with obtained interaction information, 
thus different strategies can be adopted to the fully separable or strong interaction problems. 
On the contrary, the random grouping strategy treat all problem in same way, whose performance becomes worse with the increase of dimension.
Another noteworthy phenomenon is that, the convergence rate of DA-EGO is slower than GSGA or IKAEA in early stage and is much faster in latter stage.
That's because the DA-EGO works based on the interaction analysis result, 
which is much complex in problems with higher dimension and need more accumulation of evaluated samples.
However, in the latter stage, the above obtained interaction information help the DA-EGO algorithm keep a fast convergence rate.
\begingroup
\scriptsize
\setlength{\tabcolsep}{3.3pt}
\renewcommand{\arraystretch}{1.06}
\begin{longtable}{llrrrrc}
\caption{Optimization results of 7 algorithms on the seven 60-D benchmark functions.}\label{60Dtable}\\
\toprule
Func. & Algorithm & Best & Worst & Mean & Std & Wilcoxon \\
\midrule
\endfirsthead
\multicolumn{7}{c}{Table~\ref{60Dtable} continued}\\
\toprule
Func. & Algorithm & Best & Worst & Mean & Std & Wilcoxon \\
\midrule
\endhead
\midrule
\multicolumn{7}{r}{Continued on next page}\\
\endfoot
\bottomrule
\endlastfoot
F1 & DA-EGO & 1.658E-01 & 1.109E+00 & \textbf{4.646E-01} & 2.419E-01 & N/A \\
 & RG-EGO & 6.170E-01 & 5.215E+00 & 1.849E+00 & 1.252E+00 & $+$ \\
 & DE & 5.749E+06 & 1.293E+07 & 1.017E+07 & 1.741E+06 & $+$ \\
 & GA & 2.912E+06 & 1.165E+07 & 6.508E+06 & 2.431E+06 & $+$ \\
 & GSK & 1.985E+06 & 5.321E+06 & 3.867E+06 & 9.036E+05 & $+$ \\
 & IKAEA & 2.219E+05 & 2.002E+06 & 6.489E+05 & 4.406E+05 & $+$ \\
 & GSGA & 9.081E+05 & 3.198E+06 & 1.884E+06 & 6.371E+05 & $+$ \\
\midrule
F2 & DA-EGO & 1.078E+02 & 1.853E+02 & \textbf{1.559E+02} & 1.667E+01 & N/A \\
 & RG-EGO & 1.793E+02 & 2.277E+02 & 2.098E+02 & 1.487E+01 & $+$ \\
 & DE & 6.428E+02 & 8.641E+02 & 7.802E+02 & 5.311E+01 & $+$ \\
 & GA & 3.668E+02 & 5.471E+02 & 4.680E+02 & 5.204E+01 & $+$ \\
 & GSK & 5.730E+02 & 6.953E+02 & 6.484E+02 & 2.756E+01 & $+$ \\
 & IKAEA & 3.347E+02 & 6.186E+02 & 5.304E+02 & 5.688E+01 & $+$ \\
 & GSGA & 1.643E+02 & 4.533E+02 & 3.051E+02 & 8.493E+01 & $+$ \\
\midrule
F3 & DA-EGO & 3.658E+00 & 1.153E+01 & \textbf{6.218E+00} & 1.715E+00 & N/A \\
 & RG-EGO & 9.216E+00 & 1.575E+01 & 1.271E+01 & 1.985E+00 & $+$ \\
 & DE & 1.993E+01 & 2.099E+01 & 2.057E+01 & 2.470E-01 & $+$ \\
 & GA & 1.585E+01 & 1.950E+01 & 1.799E+01 & 9.458E-01 & $+$ \\
 & GSK & 1.613E+01 & 1.857E+01 & 1.761E+01 & 6.819E-01 & $+$ \\
 & IKAEA & 3.693E+00 & 1.984E+01 & 9.455E+00 & 4.564E+00 & $+$ \\
 & GSGA & 3.939E+00 & 2.000E+01 & 8.008E+00 & 4.804E+00 & $\approx$ \\
\midrule
F4 & DA-EGO & 1.075E+05 & 1.348E+06 & \textbf{6.832E+05} & 3.276E+05 & N/A \\
 & RG-EGO & 2.587E+07 & 4.843E+07 & 3.709E+07 & 7.014E+06 & $+$ \\
 & DE & 2.065E+07 & 4.580E+07 & 3.528E+07 & 6.510E+06 & $+$ \\
 & GA & 5.075E+06 & 2.030E+07 & 1.179E+07 & 4.384E+06 & $+$ \\
 & GSK & 6.776E+06 & 1.423E+07 & 9.155E+06 & 2.018E+06 & $+$ \\
 & IKAEA & 6.459E+05 & 2.122E+06 & 1.232E+06 & 4.474E+05 & $+$ \\
 & GSGA & 9.329E+05 & 6.320E+06 & 3.028E+06 & 1.560E+06 & $+$ \\
\midrule
F5 & DA-EGO & 1.678E+02 & 2.762E+02 & \textbf{2.181E+02} & 2.816E+01 & N/A \\
 & RG-EGO & 2.632E+02 & 4.174E+02 & 3.494E+02 & 4.988E+01 & $+$ \\
 & DE & 7.363E+02 & 9.297E+02 & 8.419E+02 & 4.457E+01 & $+$ \\
 & GA & 4.312E+02 & 6.199E+02 & 5.115E+02 & 5.680E+01 & $+$ \\
 & GSK & 5.739E+02 & 7.008E+02 & 6.529E+02 & 3.520E+01 & $+$ \\
 & IKAEA & 5.002E+02 & 6.510E+02 & 5.914E+02 & 3.522E+01 & $+$ \\
 & GSGA & 1.786E+02 & 5.540E+02 & 3.190E+02 & 9.178E+01 & $+$ \\
\midrule
F6 & DA-EGO & 4.046E+01 & 8.948E+01 & \textbf{6.336E+01} & 1.436E+01 & N/A \\
 & RG-EGO & 6.543E+01 & 1.066E+02 & 8.968E+01 & 1.231E+01 & $+$ \\
 & DE & 1.225E+02 & 1.321E+02 & 1.279E+02 & 2.607E+00 & $+$ \\
 & GA & 9.156E+01 & 1.217E+02 & 1.066E+02 & 8.609E+00 & $+$ \\
 & GSK & 9.602E+01 & 1.200E+02 & 1.076E+02 & 6.172E+00 & $+$ \\
 & IKAEA & 6.397E+01 & 9.959E+01 & 8.314E+01 & 1.093E+01 & $+$ \\
 & GSGA & 6.781E+01 & 1.136E+02 & 9.316E+01 & 1.179E+01 & $+$ \\
\midrule
F7 & DA-EGO & 5.861E+03 & 2.972E+04 & 1.539E+04 & 5.032E+03 & N/A \\
 & RG-EGO & 6.643E+04 & 1.263E+05 & 6.590E+04 & 1.599E+04 & $+$ \\
 & DE & 3.474E+06 & 7.241E+06 & 5.067E+06 & 9.756E+05 & $+$ \\
 & GA & 4.631E+05 & 1.886E+06 & 9.006E+05 & 3.965E+05 & $+$ \\
 & GSK & 2.757E+05 & 9.893E+05 & 6.517E+05 & 2.124E+05 & $+$ \\
 & IKAEA & 1.401E+04 & 2.108E+05 & 6.207E+04 & 7.425E+04 & $+$ \\
 & GSGA & 2.127E+03 & 2.614E+04 & \textbf{9.931E+03} & 3.571E+03 & $-$ \\
\end{longtable}
\endgroup

\subsubsection{Comparison result of 90-D benchmark functions}
\par
\begin{figure}[htp]
  \centering
  
  \subfigure[F1]{
  \begin{minipage}[t]{0.33\linewidth}
  \centering
  \includegraphics[width=1\textwidth]{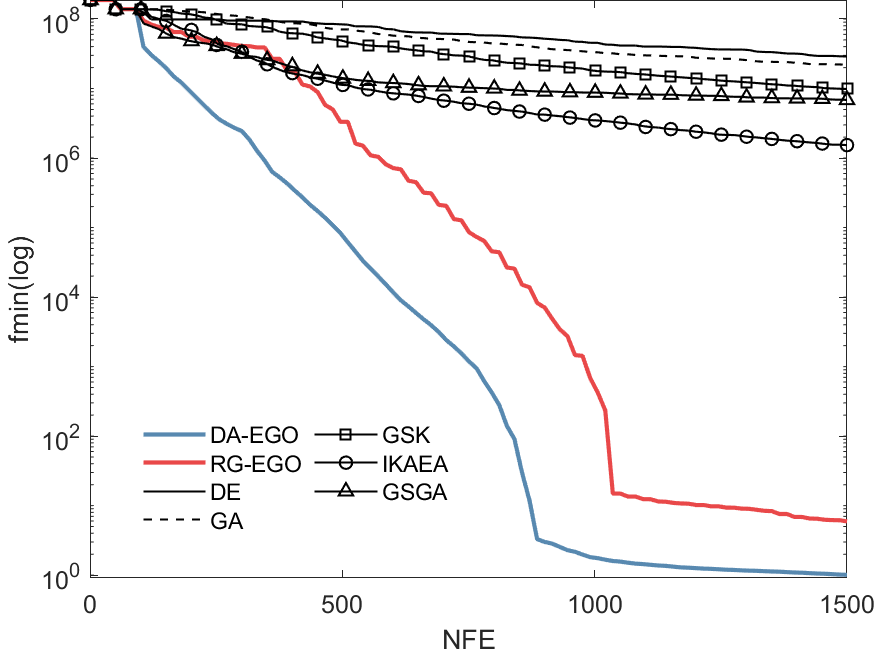}
  \end{minipage}%
  }%
  \subfigure[F2]{
  \begin{minipage}[t]{0.33\linewidth}
  \centering
  \includegraphics[width=1\textwidth]{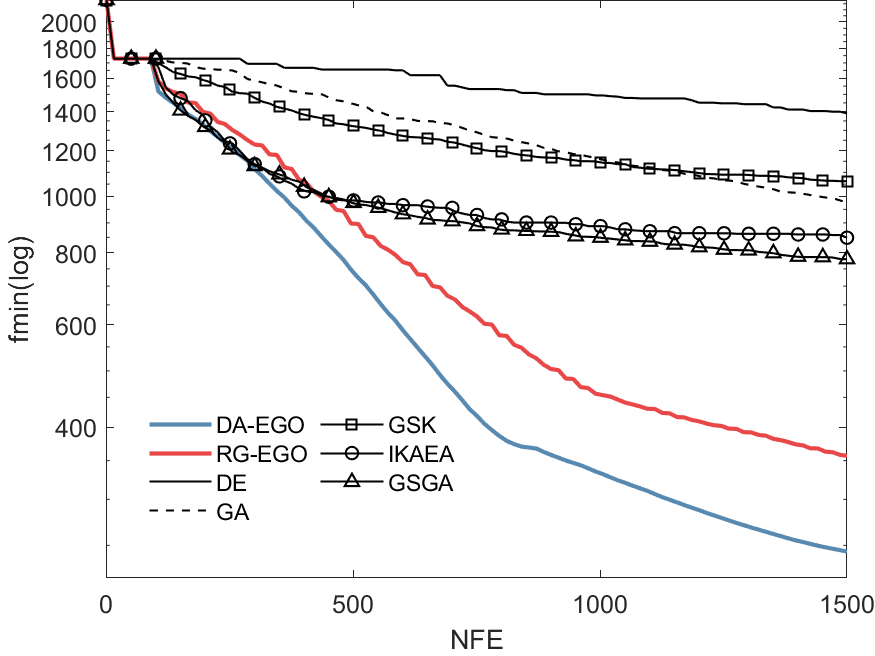}
  \end{minipage}
  }%
  \subfigure[F3]{
  \begin{minipage}[t]{0.33\linewidth}
  \centering
  \includegraphics[width=1\textwidth]{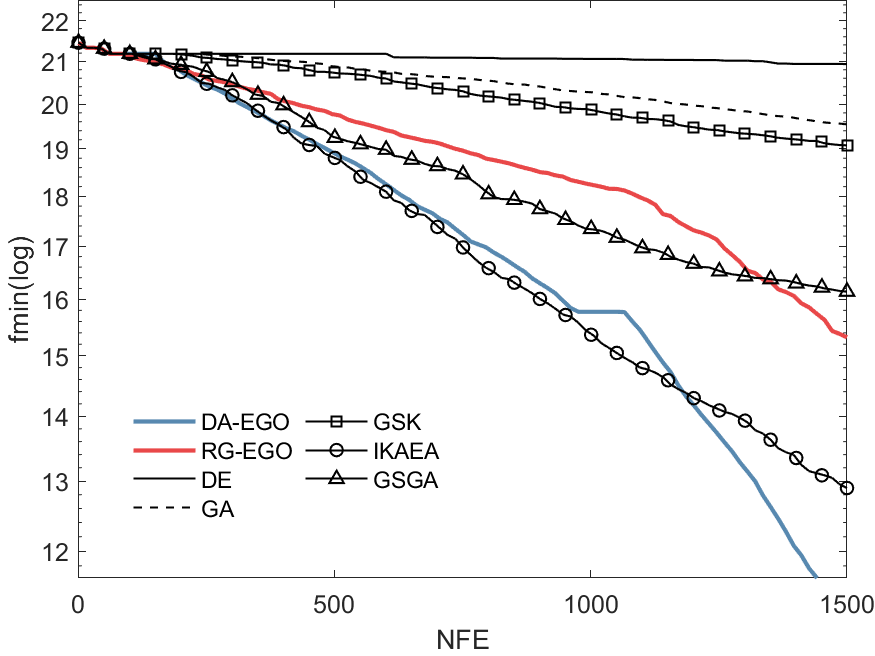}
  \end{minipage}
  }%

  \subfigure[F4]{
  \begin{minipage}[t]{0.33\linewidth}
  \centering
  \includegraphics[width=1\textwidth]{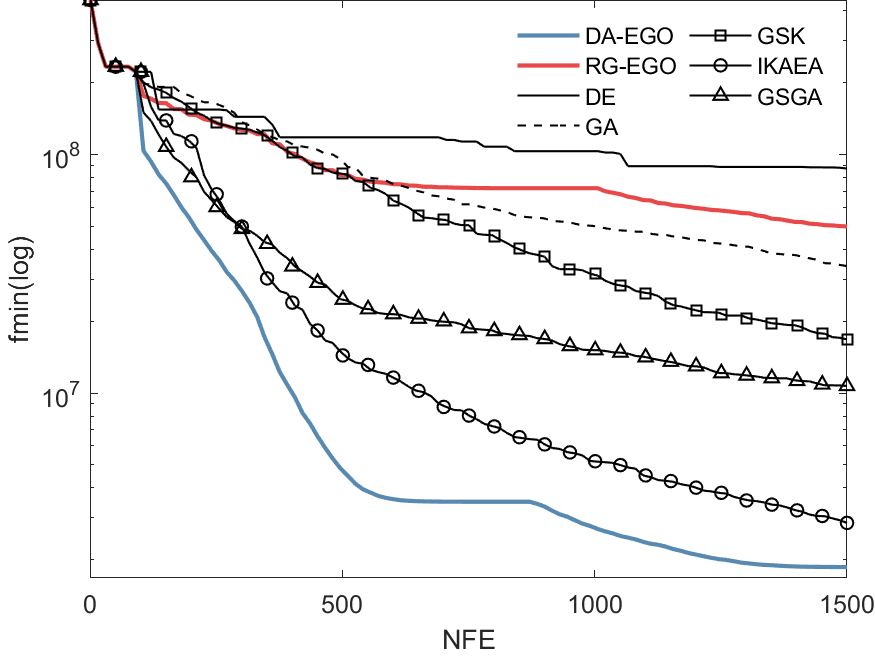}
  \end{minipage}%
  }%
  \subfigure[F5]{
  \begin{minipage}[t]{0.33\linewidth}
  \centering
  \includegraphics[width=1\textwidth]{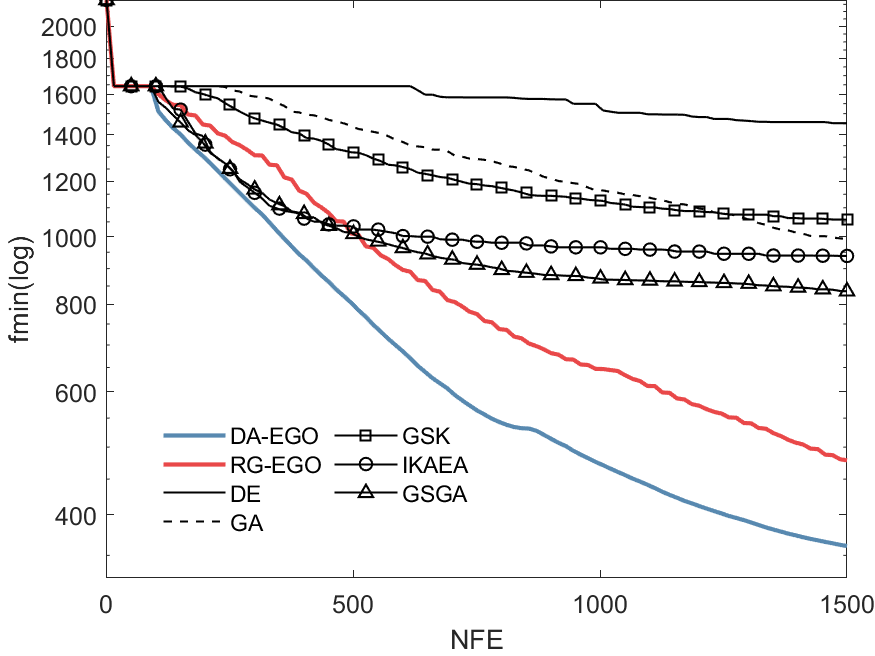}
  \end{minipage}
  }%
  \subfigure[F6]{
  \begin{minipage}[t]{0.33\linewidth}
  \centering
  \includegraphics[width=1\textwidth]{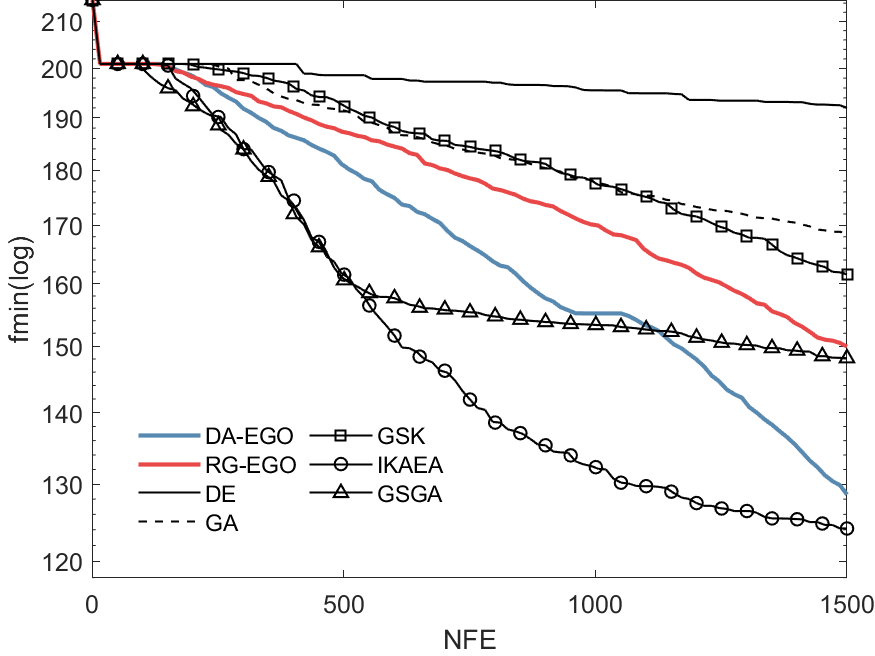}
  \end{minipage}
  }%
  
  \centering
  \caption{Average convergence curves on F1--F6 at 90 dimensions.}
  \label{90Dfigure}
  \end{figure}

\begin{figure}[htbp]
\centering
\includegraphics[width=\linewidth]{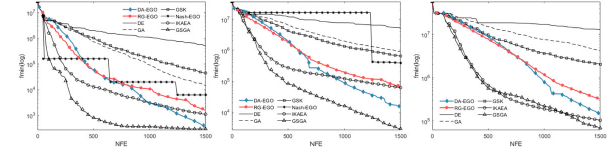}
\caption{Average convergence curves on shifted Rosenbrock (F7), from left to right: 30D, 60D, and 90D. The horizontal axis is the number of function evaluations (NFE), and the vertical axis is the objective value.}
\label{F7figure}
\end{figure}

 
The convergence curves for F1--F6 at 90 dimensions are shown in Fig.~\ref{90Dfigure}; Fig.~\ref{F7figure} shows the F7 result.
The details of the optimization results are listed in Table \ref{90Dtable}.
DA-EGO remains competitive at 90 dimensions. Its Wilcoxon comparisons with IKAEA on F3, F4, and F6 are statistically comparable, and IKAEA has the lower reported mean on F6. On F7, GSGA is significantly better than DA-EGO, while the comparison with IKAEA is statistically comparable.
\begingroup
\scriptsize
\setlength{\tabcolsep}{3.3pt}
\renewcommand{\arraystretch}{1.06}
\begin{longtable}{llrrrrc}
\caption{Optimization results of 7 algorithms on the seven 90-D benchmark functions.}\label{90Dtable}\\
\toprule
Func. & Algorithm & Best & Worst & Mean & Std & Wilcoxon \\
\midrule
\endfirsthead
\multicolumn{7}{c}{Table~\ref{90Dtable} continued}\\
\toprule
Func. & Algorithm & Best & Worst & Mean & Std & Wilcoxon \\
\midrule
\endhead
\midrule
\multicolumn{7}{r}{Continued on next page}\\
\endfoot
\bottomrule
\endlastfoot
F1 & DA-EGO & 9.531E-01 & 4.996E+00 & \textbf{2.096E+00} & 1.044E+00 & N/A \\
 & RG-EGO & 3.997E+00 & 5.521E+01 & 1.128E+01 & 9.817E+00 & $+$ \\
 & DE & 2.184E+07 & 3.440E+07 & 2.886E+07 & 4.476E+06 & $+$ \\
 & GA & 1.490E+07 & 2.899E+07 & 2.191E+07 & 4.544E+06 & $+$ \\
 & GSK & 5.962E+06 & 1.461E+07 & 9.845E+06 & 2.433E+06 & $+$ \\
 & IKAEA & 8.137E+05 & 1.954E+06 & 1.535E+06 & 4.015E+05 & $+$ \\
 & GSGA & 4.987E+06 & 9.107E+06 & 6.841E+06 & 1.429E+06 & $+$ \\
\midrule
F2 & DA-EGO & 2.530E+02 & 3.298E+02 & \textbf{2.974E+02} & 2.054E+01 & N/A \\
 & RG-EGO & 3.408E+02 & 4.434E+02 & 3.962E+02 & 2.815E+01 & $+$ \\
 & DE & 1.284E+03 & 1.498E+03 & 1.391E+03 & 6.635E+01 & $+$ \\
 & GA & 8.799E+02 & 1.094E+03 & 9.776E+02 & 6.262E+01 & $+$ \\
 & GSK & 9.438E+02 & 1.133E+03 & 1.063E+03 & 3.916E+01 & $+$ \\
 & IKAEA & 7.128E+02 & 9.533E+02 & 8.487E+02 & 7.050E+01 & $+$ \\
 & GSGA & 6.935E+02 & 8.783E+02 & 7.793E+02 & 5.303E+01 & $+$ \\
\midrule
F3 & DA-EGO & 9.354E+00 & 1.289E+01 & \textbf{1.108E+01} & 8.810E-01 & N/A \\
 & RG-EGO & 1.092E+01 & 1.871E+01 & 1.532E+01 & 2.183E+00 & $+$ \\
 & DE & 2.065E+01 & 2.106E+01 & 2.094E+01 & 1.234E-01 & $+$ \\
 & GA & 1.899E+01 & 1.993E+01 & 1.954E+01 & 3.011E-01 & $+$ \\
 & GSK & 1.836E+01 & 1.971E+01 & 1.908E+01 & 3.283E-01 & $+$ \\
 & IKAEA & 9.623E+00 & 1.944E+01 & 1.290E+01 & 2.833E+00 & $\approx$ \\
 & GSGA & 1.226E+01 & 2.035E+01 & 1.614E+01 & 3.021E+00 & $+$ \\
\midrule
F4 & DA-EGO & 6.789E+05 & 6.384E+06 & \textbf{2.657E+06} & 1.525E+06 & N/A \\
 & RG-EGO & 2.093E+06 & 1.427E+08 & 6.297E+07 & 5.888E+07 & $+$ \\
 & DE & 8.382E+07 & 1.023E+08 & 8.726E+07 & 6.402E+06 & $+$ \\
 & GA & 1.531E+07 & 5.085E+07 & 3.404E+07 & 9.625E+06 & $+$ \\
 & GSK & 1.018E+07 & 3.066E+07 & 1.678E+07 & 4.453E+06 & $+$ \\
 & IKAEA & 1.970E+06 & 4.320E+06 & 2.854E+06 & 7.711E+05 & $\approx$ \\
 & GSGA & 7.175E+06 & 1.533E+07 & 1.070E+07 & 2.634E+06 & $+$ \\
\midrule
F5 & DA-EGO & 3.599E+02 & 5.180E+02 & \textbf{4.343E+02} & 4.069E+01 & N/A \\
 & RG-EGO & 4.464E+02 & 6.829E+02 & 5.517E+02 & 5.813E+01 & $+$ \\
 & DE & 1.378E+03 & 1.554E+03 & 1.454E+03 & 6.028E+01 & $+$ \\
 & GA & 8.977E+02 & 1.060E+03 & 9.928E+02 & 4.550E+01 & $+$ \\
 & GSK & 9.815E+02 & 1.132E+03 & 1.058E+03 & 4.032E+01 & $+$ \\
 & IKAEA & 8.332E+02 & 1.040E+03 & 9.386E+02 & 6.901E+01 & $+$ \\
 & GSGA & 6.992E+02 & 8.903E+02 & 8.353E+02 & 6.896E+01 & $+$ \\
\midrule
F6 & DA-EGO & 9.898E+01 & 1.584E+02 & 1.286E+02 & 1.479E+01 & N/A \\
 & RG-EGO & 1.035E+02 & 1.708E+02 & 1.499E+02 & 1.823E+01 & $+$ \\
 & DE & 1.833E+02 & 1.976E+02 & 1.920E+02 & 3.873E+00 & $+$ \\
 & GA & 1.620E+02 & 1.787E+02 & 1.686E+02 & 6.023E+00 & $+$ \\
 & GSK & 1.456E+02 & 1.712E+02 & 1.615E+02 & 7.193E+00 & $+$ \\
 & IKAEA & 9.574E+01 & 1.513E+02 & \textbf{1.242E+02} & 1.844E+01 & $\approx$ \\
 & GSGA & 1.316E+02 & 1.611E+02 & 1.481E+02 & 7.939E+00 & $+$ \\
\midrule
F7 & DA-EGO & 5.106E+04 & 8.361E+04 & 1.418E+05 & 1.616E+04 & N/A \\
 & RG-EGO & 5.941E+05 & 9.661E+05 & 3.214E+05 & 7.055E+04 & $+$ \\
 & DE & 8.981E+06 & 1.646E+07 & 1.292E+07 & 2.854E+06 & $+$ \\
 & GA & 2.852E+06 & 5.586E+06 & 4.298E+06 & 8.603E+05 & $+$ \\
 & GSK & 1.152E+06 & 2.835E+06 & 2.048E+06 & 3.584E+05 & $+$ \\
 & IKAEA & 6.421E+04 & 2.155E+05 & 1.044E+05 & 4.341E+04 & $\approx$ \\
 & GSGA & 5.116E+04 & 1.150E+05 & \textbf{9.871E+04} & 1.920E+04 & $-$ \\
\end{longtable}
\endgroup

\par
Across the 21 instances, DA-EGO has the lowest reported mean on 17 instances and is competitive on several others. The interaction-based decomposition is effective on the separable and partially separable functions, but it does not give uniformly superior results on the non-separable shifted Rosenbrock function. In particular, the 60-D and 90-D F7 Wilcoxon comparisons favor GSGA. These results support DA-EGO as an approach for expensive high-dimensional optimization, with performance depending on the interaction structure of the problem.

\subsubsection{Computational time}
Table~\ref{runtimeTable} reports the average runtime and mean optimum for the shifted elliptic function (F1). DA-EGO takes 35, 36, and 40 minutes at 30, 60, and 90 dimensions, respectively. Model-free algorithms have substantially lower computational overhead; IKAEA also reduces model-building time using its enhanced Kriging implementation. The runtimes of DA-EGO, RG-EGO, GSGA, and Nash-EGO are of the same order. DA-EGO therefore trades additional surrogate construction and interaction analysis for fewer expensive function evaluations. This trade-off is most relevant when evaluating the physical objective dominates the cost of the optimization algorithm.
\begin{table}[htbp]
\centering
\caption{Average runtime and mean optimum on shifted elliptic (F1) at 30, 60, and 90 dimensions. Time units are shown explicitly.}
\label{runtimeTable}
\small
\begin{tabular}{llrr}
\toprule
Dimension & Algorithm & Average time & Mean optimum \\
\midrule
30D & DA-EGO & 35 min & 1.73E-02 \\
 & RG-EGO & 36 min & 3.30E-01 \\
 & DE & 0.741 s & 1.34E+06 \\
 & GA & 3.2 s & 1.81E+06 \\
 & GSK & 0.124 s & 4.59E+05 \\
 & Nash-EGO & 46 min & 2.44E+02 \\
 & IKAEA & 68 s & 8.11E+04 \\
 & GSGA & 26 min & 2.64E+05 \\
\midrule
60D & DA-EGO & 36 min & 4.65E-01 \\
 & RG-EGO & 33 min & 1.85E+00 \\
 & DE & 0.919 s & 1.02E+07 \\
 & GA & 3.31 s & 6.51E+06 \\
 & GSK & 0.139 s & 3.87E+06 \\
 & IKAEA & 101 s & 6.49E+05 \\
 & GSGA & 30 min & 1.88E+06 \\
\midrule
90D & DA-EGO & 40 min & 2.10E+00 \\
 & RG-EGO & 39 min & 1.13E+01 \\
 & DE & 1.264 s & 2.89E+07 \\
 & GA & 3.65 s & 2.19E+07 \\
 & GSK & 0.194 s & 9.85E+06 \\
 & IKAEA & 149 s & 1.54E+06 \\
 & GSGA & 37 min & 6.84E+06 \\
\bottomrule
\end{tabular}
\end{table}

\subsection{Parameter sensitivity analysis}
\begin{figure}[htbp]
\centering
\includegraphics[width=\linewidth]{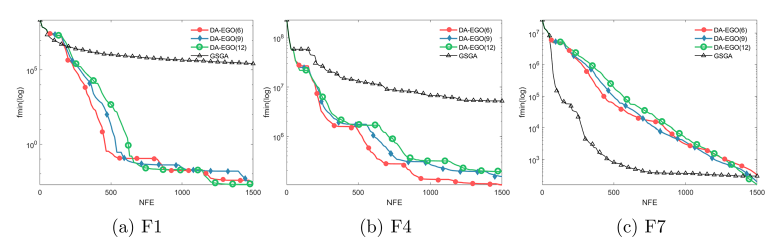}
\caption{Average convergence curves for sub-task iteration multipliers $t_{iter}=6,9,12$ on (a) F1, (b) F4, and (c) F7. Numbers in the DA-EGO legend indicate $t_{iter}$; GSGA is included for comparison. NFE denotes the number of function evaluations.}
\label{parameterSensitivity}
\end{figure}

The maximum number of sub-task iterations is written as $iter_{\max}=t_{iter}d$, where $d$ is the subspace dimension. The default is $t_{iter}=6$, while the sensitivity study compares $t_{iter}\in\{6,9,12\}$. Figure~\ref{parameterSensitivity} shows the average convergence curves on F1, F4, and F7, with GSGA included as a comparator. Within this tested range, changes in $t_{iter}$ have a limited effect on the overall DA-EGO convergence trend. Shorter sub-task runs can make interaction information available to the next decomposition cycle earlier. This observation is limited to the three functions and parameter values shown.
\section{Engineering Test Case}
As mentioned in the introduction, the design of turbomachinery components are typical HEB problems, 
especially the joint design of multiple turbomachinery components.
After using benchmark functions to prove the efficiency and robustness of the DA-EGO, 
in this section, two engineering optimization design tasks will be solved by the proposed algorithm. 
\par 
The compressor is a turbomachine that converts mechanical energy into fluid kinetic energy and potential energy. 
As an important part of a gas turbine, the performance of the compressor has a direct impact on the efficiency, power, and reliability of the gas turbine. 
The optimization design of the compressor is a difficult task. 
Due to the strong three-dimensional effect of the flow in the compressor, 
the shape of the compressor blade is usually very complex, and the profile shape from the root to the tip varies greatly, with obvious bending and sweeping at the same time.
Therefore, more design variables are needed to accurately shape the compressor blade. 
\par
Moreover, in a multistage axial flow compressor, each row of blades is disturbed by upstream and downstream blades. 
So, the matching between different blades has a great impact on the efficiency of the compressor, 
and the multistage blades should to be considered as a whole in the design.
However, multistage compressor design brings two problems: 
(\romannumeral1) the expansion of the calculation domain leads to the increase of the number of grids, 
and usually the computational cost of each performance evaluation increases linearly with the number of blades; 
(\romannumeral2) the increase of the number of blades leads to the increase of design variables. 
Usually, the scale of problem-solving increases exponentially with the number of design variables. 
\par
In this section, a single compressor design task with 28 design variables and a multistage compressor design task with 60 variables are completed with the proposed DA-EGO algorithm.
And the DA-EGO's performance are compared with other four algorithms, with tha same initial distribution of 50 samples that generated by the Matlab built-in function ``lhsdesign". 
\subsection{Engineering test case 1: Rotor 37 blade}
\par 
In the first engineering test case, the proposed DA-EGO optimization algorithm is applied to the aerodynamic optimization of the well-known Rotor 37 blade~\citep{suderExperimentalInvestigationFlow1996}. 

\subsubsection{Problem description}
\par
\begin{figure}[ht]
\begin{center}
\includegraphics[width=1\textwidth, trim = 0 0.5cm 0 0.5cm]{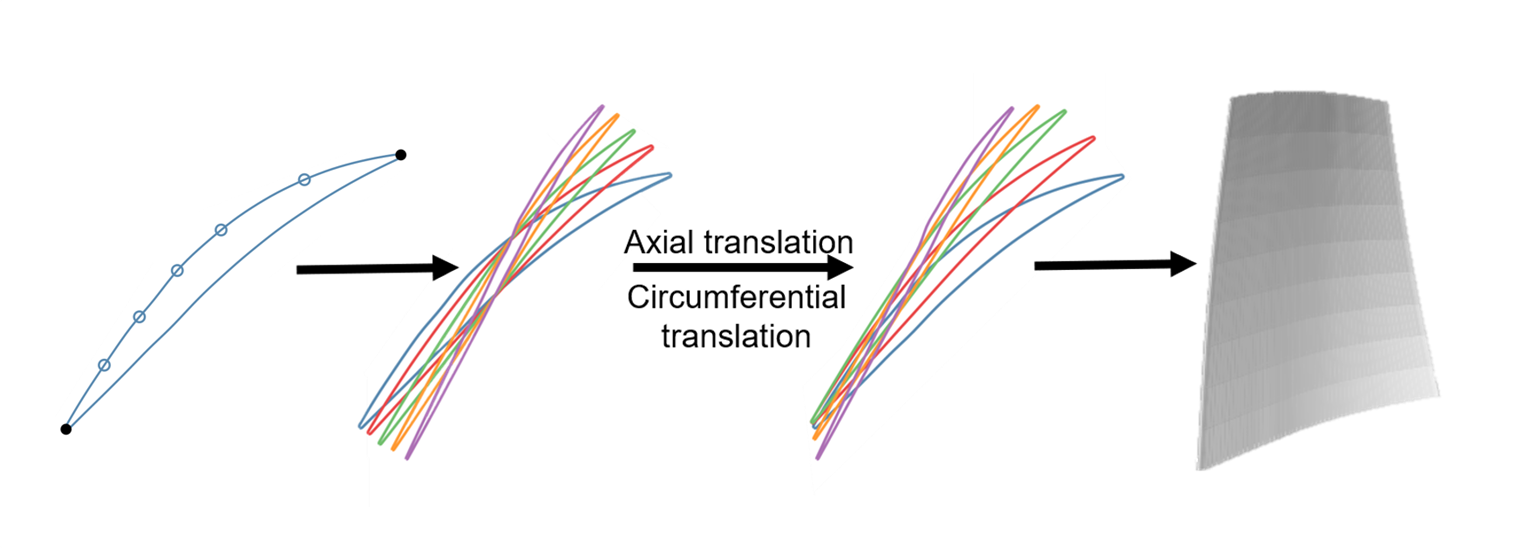}
\end{center}
\caption{ The sketch map of the 3D parameterization method}  
\label{bladeconstruct} 
\end{figure}

Here we select 5 section profiles of 0\%, 25\%, 50\%, 75\%, and 100\% span, and  5 active control points at the suction side of each section are selected to adjust the section profiles. 
At the same time, the bending and sweeping of the blade are also adjusted. 
The number of design variables in the optimization of Rotor 37 is 28, these variables are shown in Table \ref{Rotor37Variable}.
Figure \ref{bladeconstruct} shows the generation of three-dimensional blade geometric modeling. 
The parameters of the 5 control points of each section determine the shape of this section. 
After finishing the parameterization of all section profiles, 1 circumferential translation parameter $x_{26}$ for the middle section, and 2 axial translation parameters $x_{27},x_{28}$ for the tip section and middle section are selected to adjust the stacking line in 3D space.
Then, a 3D blade profile is obtained using skinning surface techniques.
\begin{table}[htbp]
  \centering
  \caption{Design variables in engineering test case}
    \begin{tabular}{cc}
    \toprule
   Geometric definition & variable index \\   
        \midrule
    control coefficient in 0\% span & $x_1,\cdots,x_5$ \\
    control coefficient in 25\% span &$x_6,\cdots,x_{10}$  \\
    control coefficient in 50\% span & $x_{11},\cdots,x_{15}$  \\
    control coefficient in 75\% span &$x_{16},\cdots,x_{20}$\\
    control coefficient in 100\% span & $x_{21},\cdots,x_{25}$\\
    circumferential translation parameter & $x_{26}$  \\
    axial translation parameter & $x_{27},x_{28}$  \\
        \bottomrule
    \end{tabular}%
  \label{Rotor37Variable}%
\end{table}%
The isentropic efficiency $\eta _{is}$ is set as the objective function for the optimization, the related optimization model is shown below:
\begin{equation}
\begin{array}{c}
\max  \{f_{obj}(\mathbf{x})\}=\max\{\eta_{is}(\mathbf{x})/p(\mathbf{x})\} \\
\text { s.t. }  0.98 \cdot m(\text {ref}) \leq m(\mathbf{x}) \leq 1.02 \cdot m(\text {ref}) \\
\qquad 0.98 \cdot Pr(\text {ref}) \leq Pr(\mathbf{x}) \leq 1.02 \cdot Pr(\text {ref}) \\
\text{where    }{\eta _{is}} = 
{
\{\left({p_{outlet}^t}/{p_{inlet}^t}\right)
^\frac{\gamma-1}{\gamma} - 1}
\}/
{
\left({{T_{outlet}^t}/{T_{inlet}^t} - 1}\right)
}
\end{array}
\end{equation}
where,  $\text {ref}$ means the reference design, $m$ is the mass flow rate, $Pr = {p_{outlet}^t}/{p_{inlet}^t}$ is the total pressure ratio, and $\gamma$ denotes adiabatic exponent.
In the meantime, the design point flow of the optimized design is constrained so that its change does not exceed 2\% of the reference design mass flow. The constraint is realized in the form of penalty function $p(x)$ ~\citep{songResearchMetamodelBasedGlobal2016}.

\subsubsection{Numerical simulation model}
\par
\begin{figure}[htbp]
  \centering
  \subfigure[efficiency-mass flow]{
  \begin{minipage}[t]{0.45\linewidth}
  \centering
  \includegraphics[width=1\textwidth]{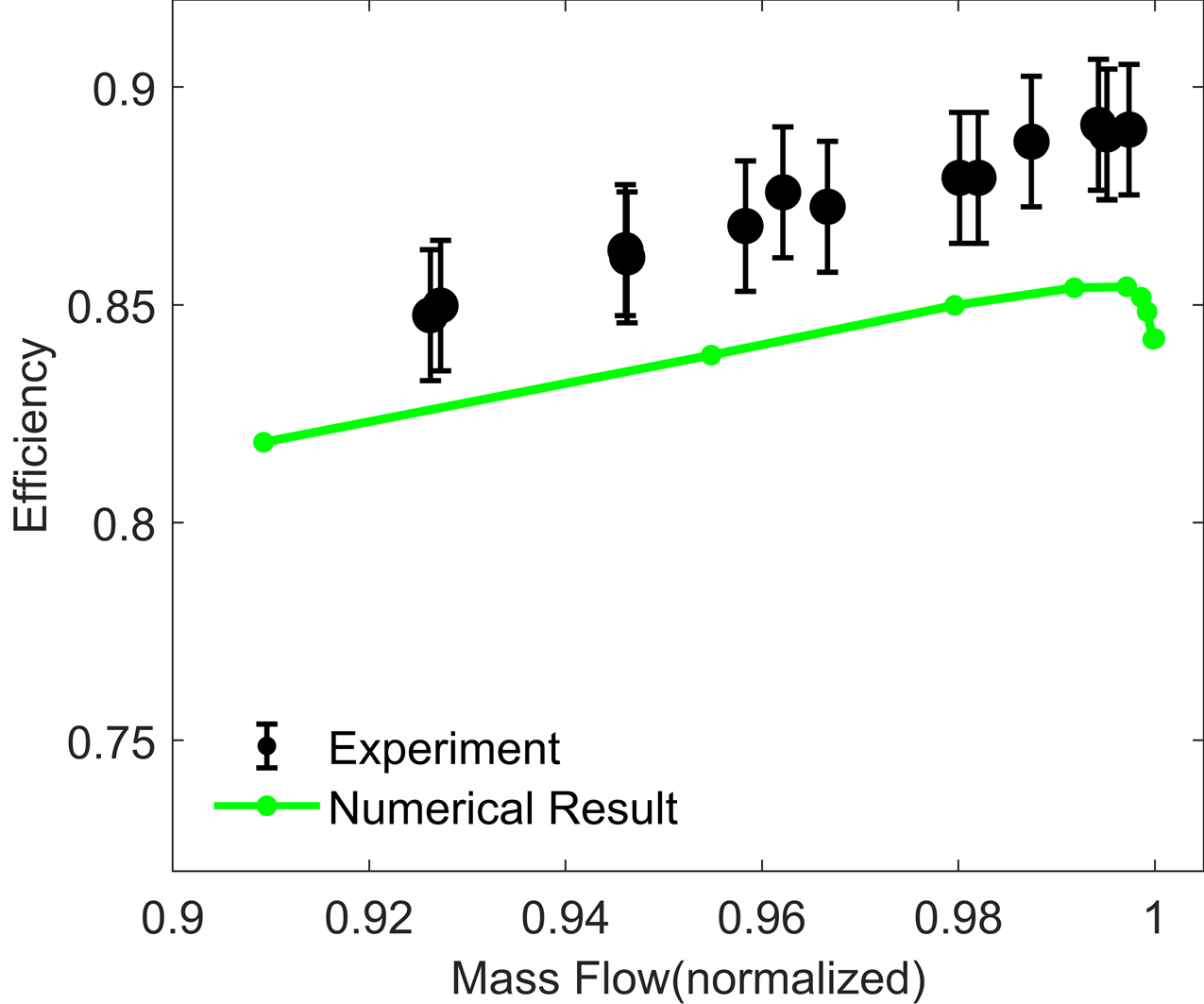}
  \end{minipage}%
  }%
  \subfigure[pressure ratio-mass flow]{
  \begin{minipage}[t]{0.45\linewidth}
  \centering
  \includegraphics[width=1\textwidth]{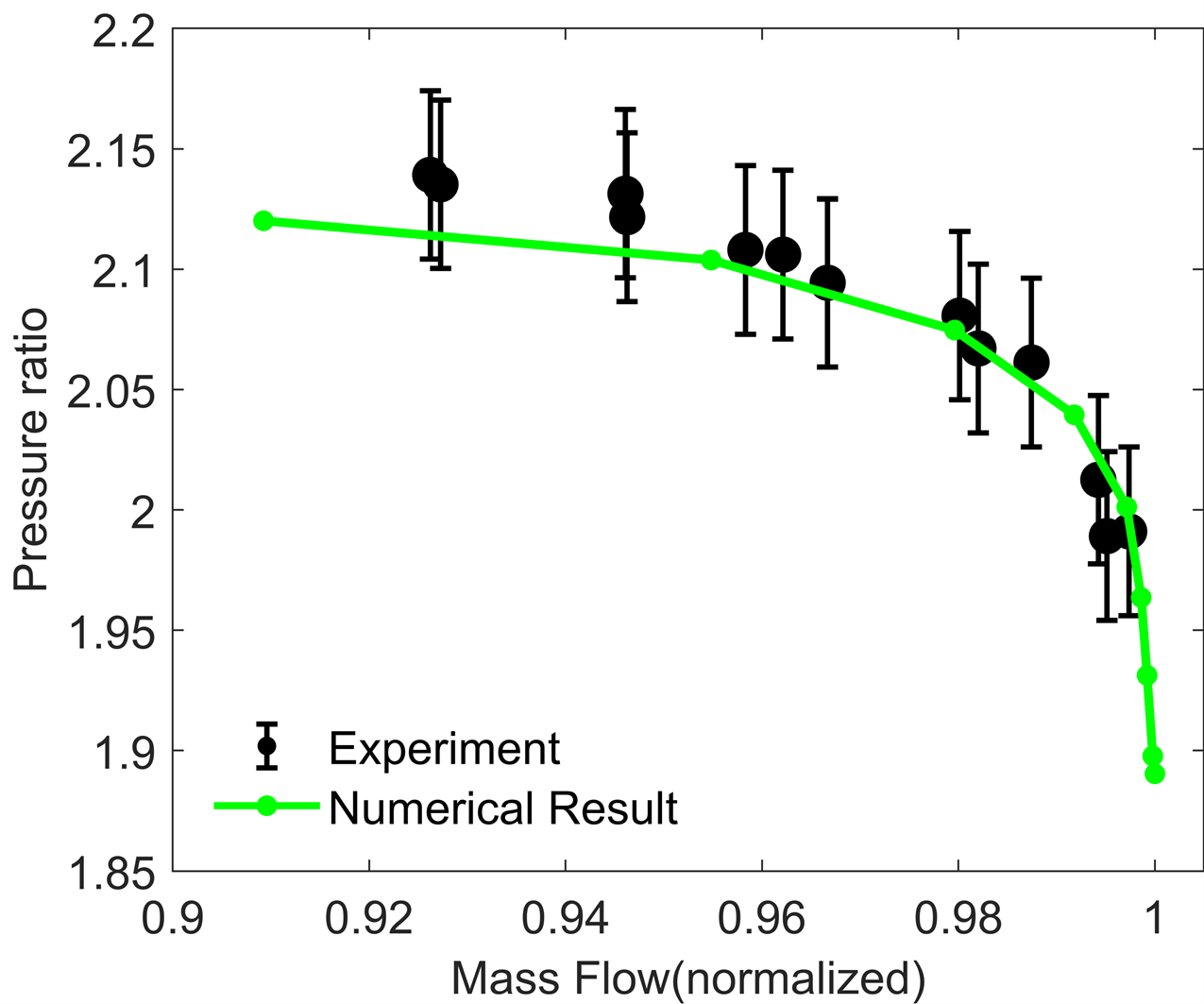}
  \end{minipage}
  }%
  \centering
  \caption{The comparison of feature curves of CFD simulation and experimrnt result}
  \label{Rotor37EXP}
  \end{figure}

The Rotor 37 blade is one of the rotors of the four-stage axial flow compressor Stage 37 with a high-pressure ratio. It was designed and tested by Reid and Moore in the NASA Glenn center in the 1970s. The design parameters are taken as the inlet stage parameters of a typical aero-engine. 
Table \ref{designConditionRotor37} shows the relevant design parameters of Rotor 37, which keep the same as the literature~\cite{borettiExperimentalComputationalAnalysis2010}. 
To be in accordance with the literature, uniform total pressure and temperature are imposed at the inlet boundary, and an averaged static pressure is imposed at the outlet. The optimization is carried out with a constant outlet static pressure of 115000 Pa, corresponding to a relative mass flow rate of 99\% for the Rotor 37 blade.
\begin{table}[htbp]
  \centering
  \caption{Design conditions of the Rotor 37 blade}
    \begin{tabular}{cc}
    \toprule
    Condition name & Value \\
    \midrule
    equivalent rotational speed[rpm] & 17188.7 \\
    number of rotor blades & 36 \\
    rotor blade aspect ratio & 1.19 \\
    tip clearance gap[mm] & 0.356 \\
    inlet total temperature[K] & 288.15 \\
    inlet total pressure[Pa] & 101325 \\
    \bottomrule
    \end{tabular}%
  \label{designConditionRotor37}%
\end{table}%
For the above model, a grid with about  $4 \times {10^{ 6}}$ nodes is used to calculate. 
The H–O–I topology is employed for the generation of the grid by using the commercial software NUMECA auto-grid5, and the mesh thickness of the first layer near the wall is set to be $3 \times {10^{ - 6}}$ m. 
In the calculation, the Spalart-Allmaras turbulence model is used with adiabatic smooth walls condition. 
The Reynolds-averaged Navier-Stokes equations are solved by using the commercial software NUMECA FINE/TURBO. 
With the CPU Intel(R) i5-9400F@2.9GHz, the calculation time of a single sample is about 600s. 
In the design condition, the efficiency, pressure ratio, and mass flow of the reference design are 85.39\%, 2.0395, and 20.62kg/s respectively.
Figure \ref{Rotor37EXP} shows the comparison between the calculated results and the experimental results. It is easy to see that the calculated results of CFD are in good agreement with the experimental results, which verifies the correctness of the CFD calculation method.
\subsubsection{Results Analysis}
\par
These algorithms are used to optimize the Rotor 37 blade 5 times independently, and the total number of CFD calculations in each optimization is 1000. 
The optimization results are shown in Table \ref{Rotor371Result}. 
All 5 algorithms improve the efficiency of the blade, among which the optimization result of the DA-EGO algorithm is the best, which has great advantages in convergence efficiency and robustness compared with other algorithms. 
After 1000 times of CFD calculations with the DA-EGO, the total efficiency of Rotor 37 blade has been improved by 1.65\%. The variance of the results of 5 repeated operations of DA-EGO, GA, and DE algorithms is very small, indicating that these algorithms are less affected by the initial distribution of samples.
\begin{table*}[htbp]
  \centering
\caption{The optimization results of the engineering test case 1 in detail.}
\resizebox{\linewidth}{!}{
\begin{tabular}{lcccc}
\toprule
    Algorithm & Efficiency  &Pressure ratio &Mass flow(kg/s) &\makecell[c]{Efficiency\\ improvement}\\
    \midrule
    Baseline    & 0.8539  & 2.040  & 20.62  & --- \\
    DA-EGO    & 0.8707  & 2.055  & 20.96  & 1.65\% \\
    GSGA  & 0.8696  & 2.051  & 20.89  & 1.57\% \\
    IKAEA & 0.8665  & 2.050  & 20.88  & 1.26\% \\
    GA    & 0.8673  & 2.049  & 20.88  & 1.34\% \\
    DE    & 0.8645  & 2.049  & 20.83  & 1.06\% \\
    \bottomrule
    \end{tabular}}%
  \label{Rotor371Result}%
\end{table*}%
\par
\begin{figure}[ht]
\centering
\subfigure[baseline]{
\begin{minipage}[t]{0.5\linewidth}
\centering
\includegraphics[scale=0.3,trim=8cm 0.5cm 8cm 0.5cm,clip]{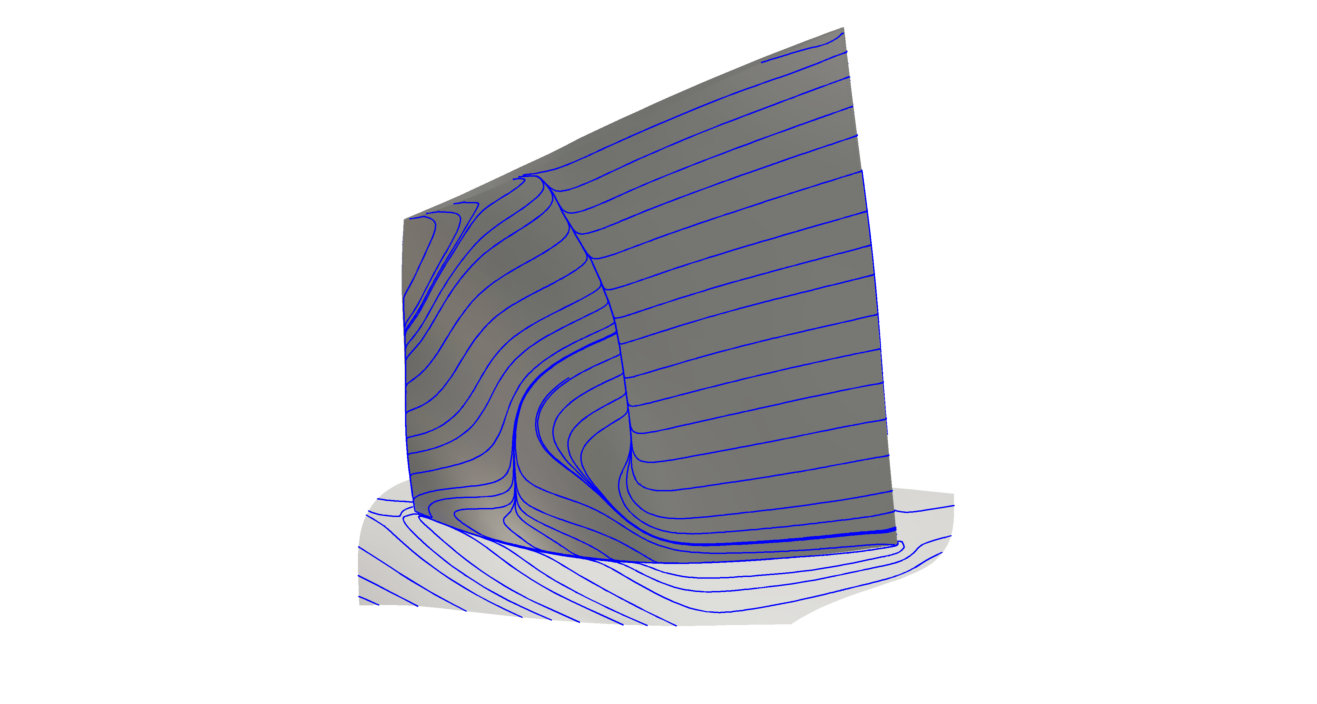}
\end{minipage}%
}%
\subfigure[DA-EGO]{
\begin{minipage}[t]{0.5\linewidth}
\centering
\includegraphics[scale=0.3,trim=8cm 0.5cm 8cm 0.5cm,clip]{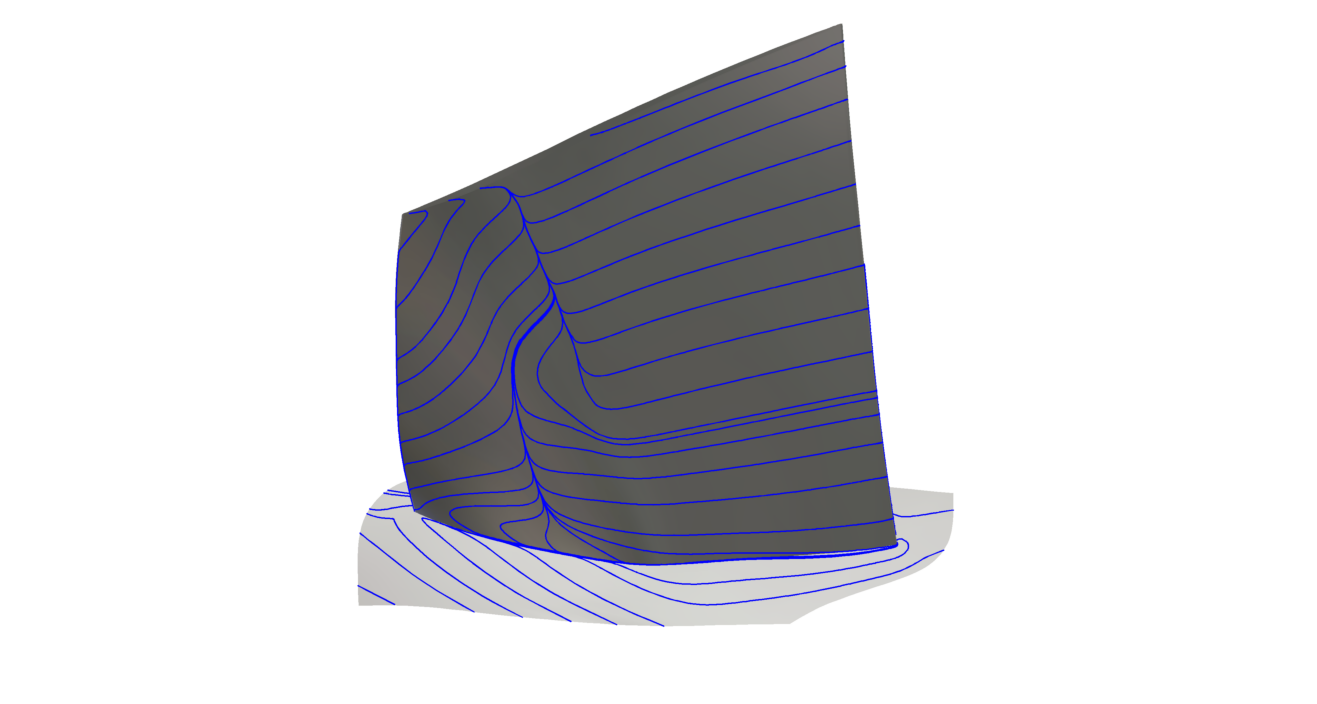}
\end{minipage}
}%

\centering
\caption{The limit streamlines the suction surface of the DA-EGO algorithms and the baseline design of the Rotor 37 blade.}
\label{Rotor37stream}
\end{figure}

Figure \ref{Rotor37stream} exhibits the suction surface's limited flow for the baseline design and optimization results from the DA-EGO algorithm. 
This comparison reveals the presence of the ``shock wave-boundary layer interference" phenomenon on the baseline Rotor 37 suction surface, and severe flow separation near the trailing edge. 
Furthermore, there is involved back-flow near the trailing tip edge. 
Comparing Fig.\ref{Rotor37stream}(a) to Fig.\ref{Rotor37stream}(b), highly-optimized separation lines resulted in reduced areas of separation and intensity of involved back-flow, which is beneficial to the improvement of efficiency. 

\subsection{Engineering test case 2: Multi-stage compressor}
\subsubsection{Problem description}
\begin{figure}[ht]
  \centering
  \subfigure[Geometry]{
    \begin{minipage}[t]{0.45\linewidth}
    \centering
    \includegraphics[width=1\textwidth]{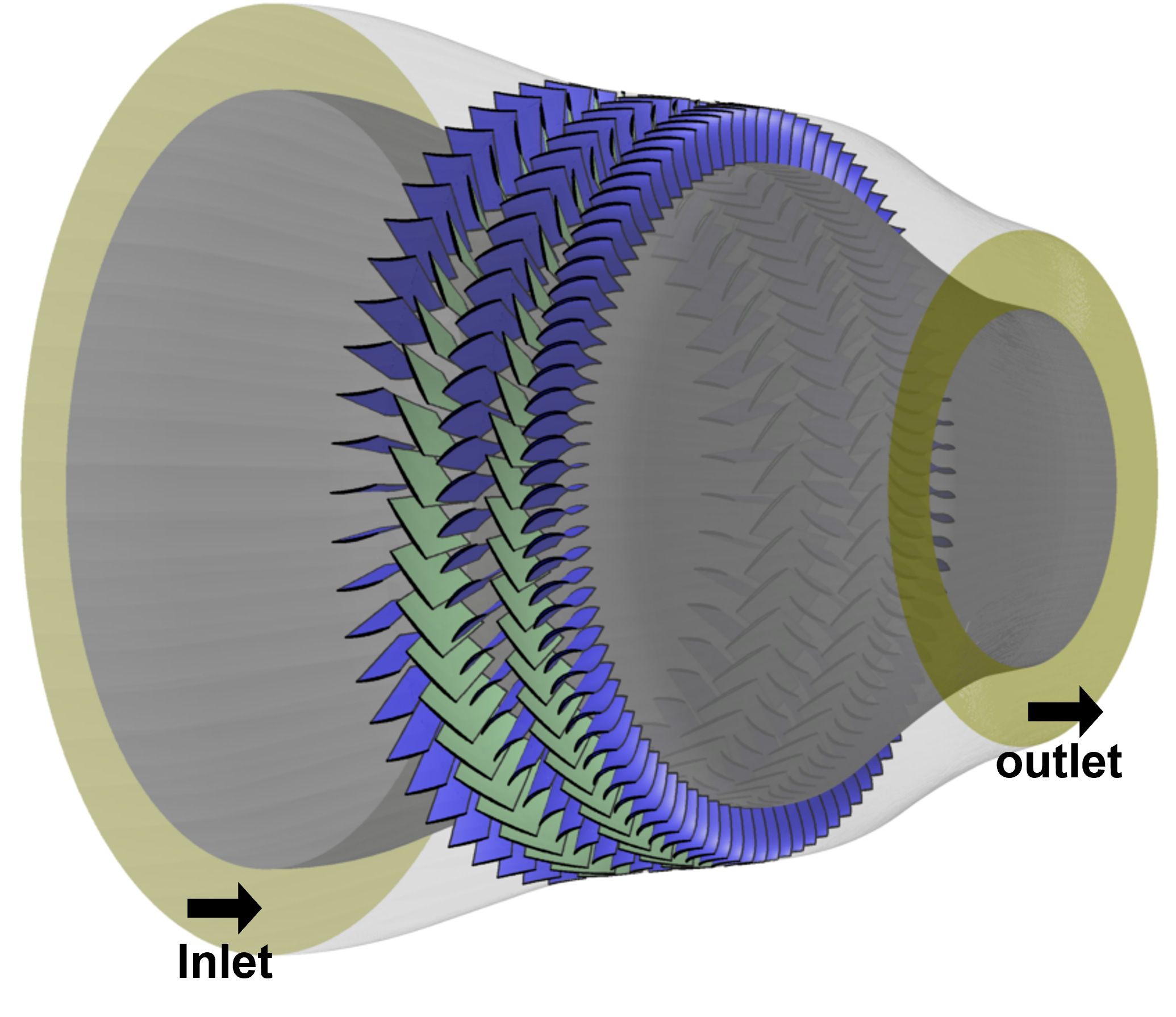}
    \end{minipage}%
    }%
  \subfigure[Mesh]{
    \begin{minipage}[t]{0.55\linewidth}
    \centering
    \includegraphics[width=1\textwidth]{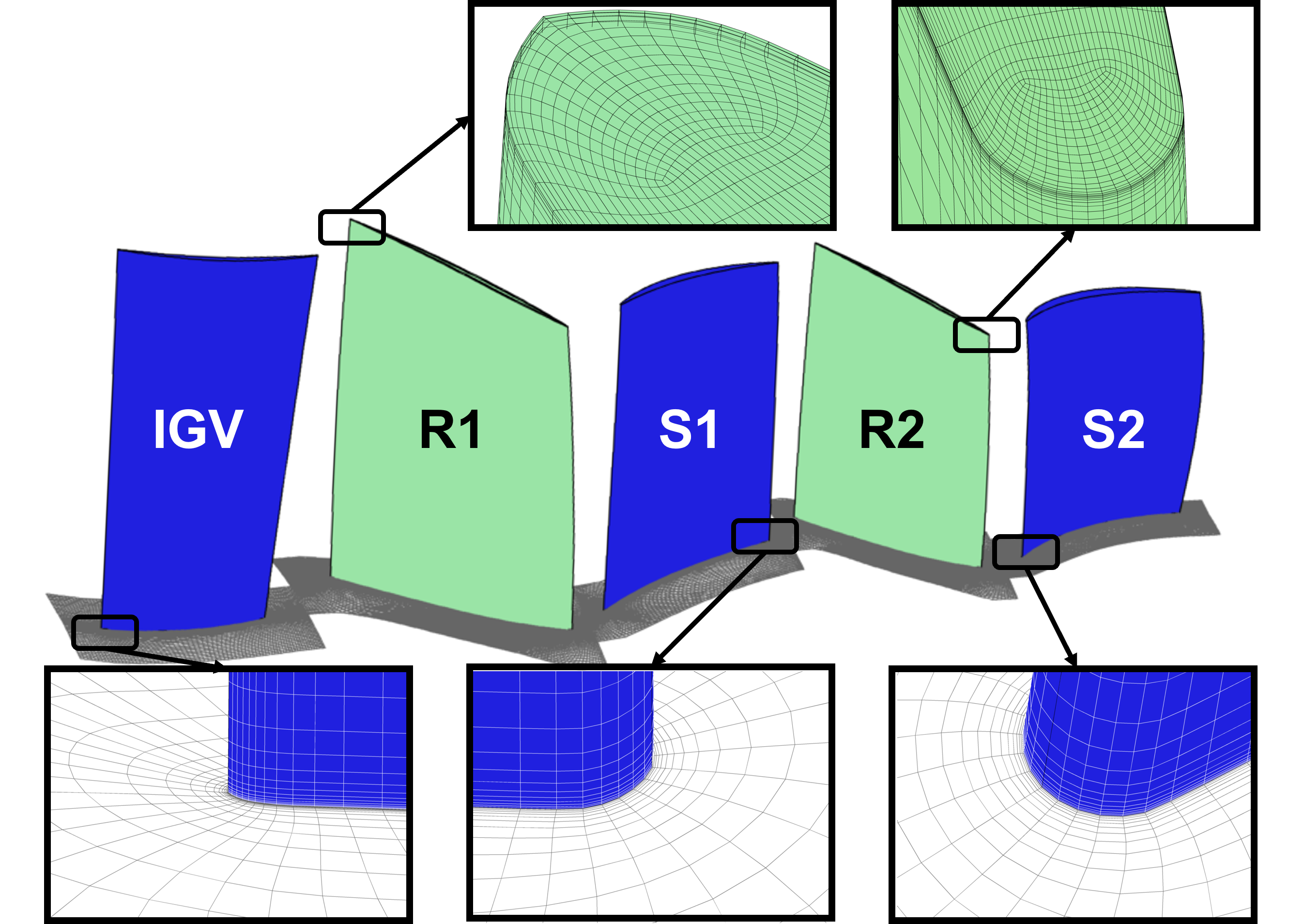}
    \end{minipage}%
    }%
\centering
\caption{The geometry model and mesh of the multistage compressor.}   
\label{GTFgeom}
\end{figure}

In this section, a typical multistage axial flow compressor design task is used to demonstrate the efficiency of the DA-EGO algorithm in engineering tasks with more design variables.
The geometry and mesh of the design object are shown in Fig.\ref{GTFgeom}.
This multistage axial compressor is consisted with five rows of blades, 
which contain one inlet guide vane (IGV) and two compressor stages.
In this case, the grid used in optimization with about  $2.3 \times {10^{ 7}}$ nodes is also generated by the software Autogrid5. 
With the CPU Intel(R) i5-9400F@2.9GHz, the calculation time of a single sample of the multi-stage compressor is about 4200s. 
Table \ref{designConditionGTF} shows the relevant design parameters of the above multistage axial compressor.
\begin{table}[htbp]
  \centering
  \caption{Design conditions of the multistage compressor}
    \begin{tabular}{cc}
    \toprule
    Condition name & Value \\
    \midrule
    equivalent rotational speed[rpm] & 8614.2 \\
    number of rotor blades & 51/47/67/57/97 \\
    tip clearance gap[mm] & 0.3 \\
    inlet total temperature[K] & 288.15 \\
    inlet total pressure[Pa] & 101325 \\
    design average outlet static pressure[Pa] & 220000\\
    design total pressure ratio & 2.34\\
    \bottomrule
    \end{tabular}%
  \label{designConditionGTF}%
\end{table}%
\par
Similar to the previous engineering test case, the parameterization method of each blade is shown in Fig.\ref{bladeconstruct}.
In each stage, whose rotor blade and stator are labeled as R1/R2 and S1/S2,
three section profiles of 0\%, 50\%, and 100\% span are selected from each blade.
Considering that the flow in the rotor blade's passage is more complex, 
five active control points are selected from each section profile's suction curve in rotor blade,
and only three control points are selected in the stator's suction.
In addition to the three variables that control the bending and sweeping, finally, the rotor blade R1/R2 has 18 design variables and the stator blade S1/S2 has 12 design variables.
As the IGV blade remains the same in this optimization design, it totally has 60 design variables for the remaining four rows of blades.
\subsubsection{ Result Analysis}
\par
The DA-EGO and the other four compared algorithms are used to optimize the multistage axial compressor in one time, and the total number of CFD calculations in each optimization is also 1000.
The optimization results are shown in Table \ref{Engineer1ResultGTF}.
\begin{table*}[htbp]
  \centering
\caption{The optimization results of the engineering test case 2 of the multistage axial compressor in detail.}
\begin{tabular}{lcccc}
\toprule
    Algorithm & Efficiency  &Pressure ratio &Mass flow(kg/s) &\makecell[c]{Efficiency\\ improvement}\\
    \midrule
    Baseline & 88.072\% & 2.4604 & 53.6465 & --- \\
    DA-EGO & 89.142\% & 2.4644 & 53.9005 & 1.070\% \\
    GSGA  & 88.973\% & 2.4630 & 53.7780 & 0.901\% \\
    IKAEA & 88.905\% & 2.4623 & 53.8325 & 0.833\% \\
    GA    & 88.804\% & 2.4634 & 53.8110 & 0.732\% \\
    DE    & 88.495\% & 2.4575 & 53.5025 & 0.423\% \\
    \bottomrule
    \end{tabular}%
  \label{Engineer1ResultGTF}%
\end{table*}%
It's obvious that the DA-EGO algorithm is significantly better than all other compared algorithms on the test of multistage compressor design optimization.
The proposed DA-EGO method improves the total efficiency of the multistage compressor by 1.07\%.
\par
\begin{figure}[htbp]
\centering
\subfigure[the whole compressor]{
\begin{minipage}[t]{0.33\linewidth}
\centering
\includegraphics[width=0.7\textwidth]{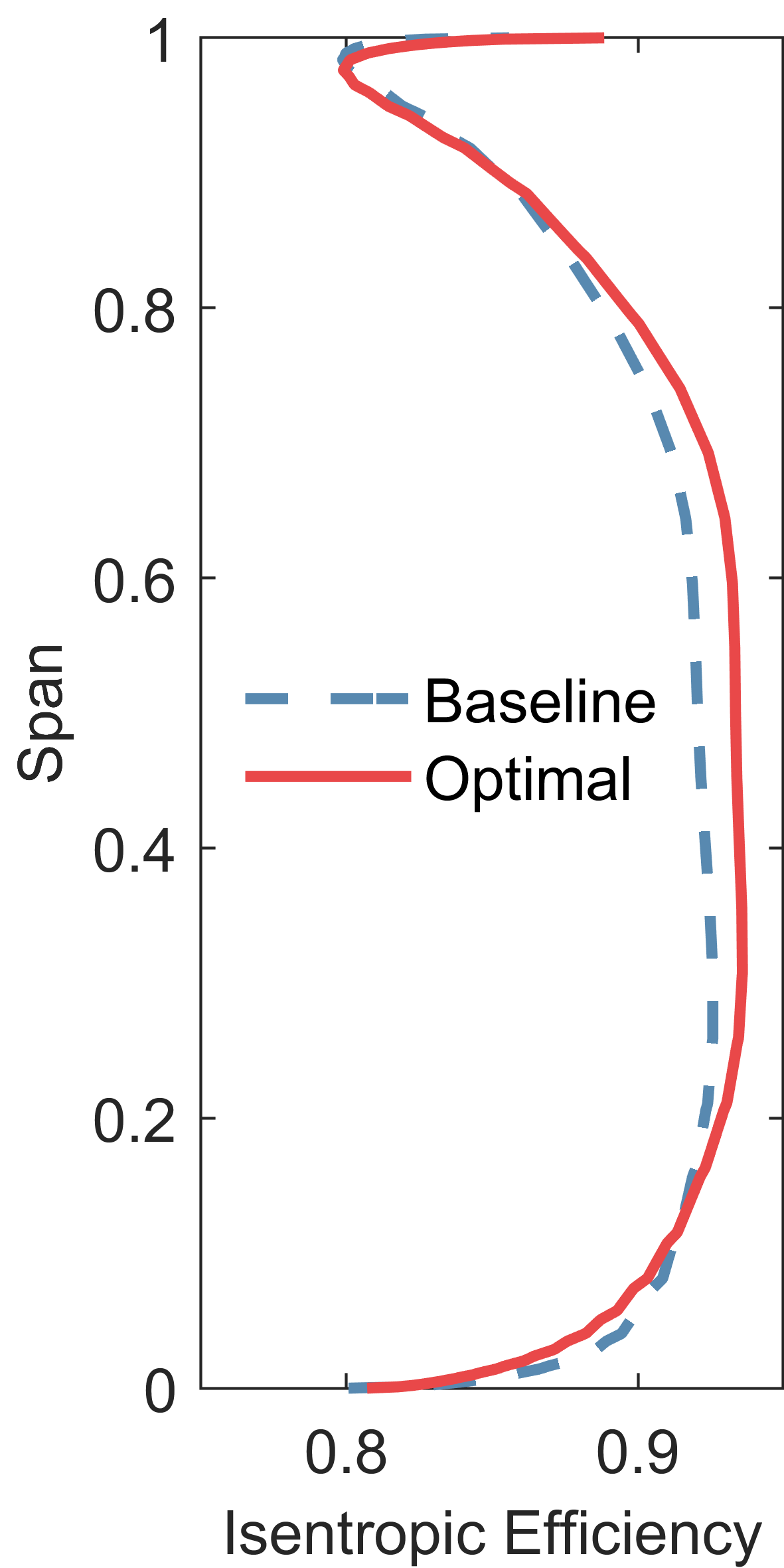}
\end{minipage}%
}%
\subfigure[the first stage]{
\begin{minipage}[t]{0.33\linewidth}
\centering
\includegraphics[width=0.7\textwidth]{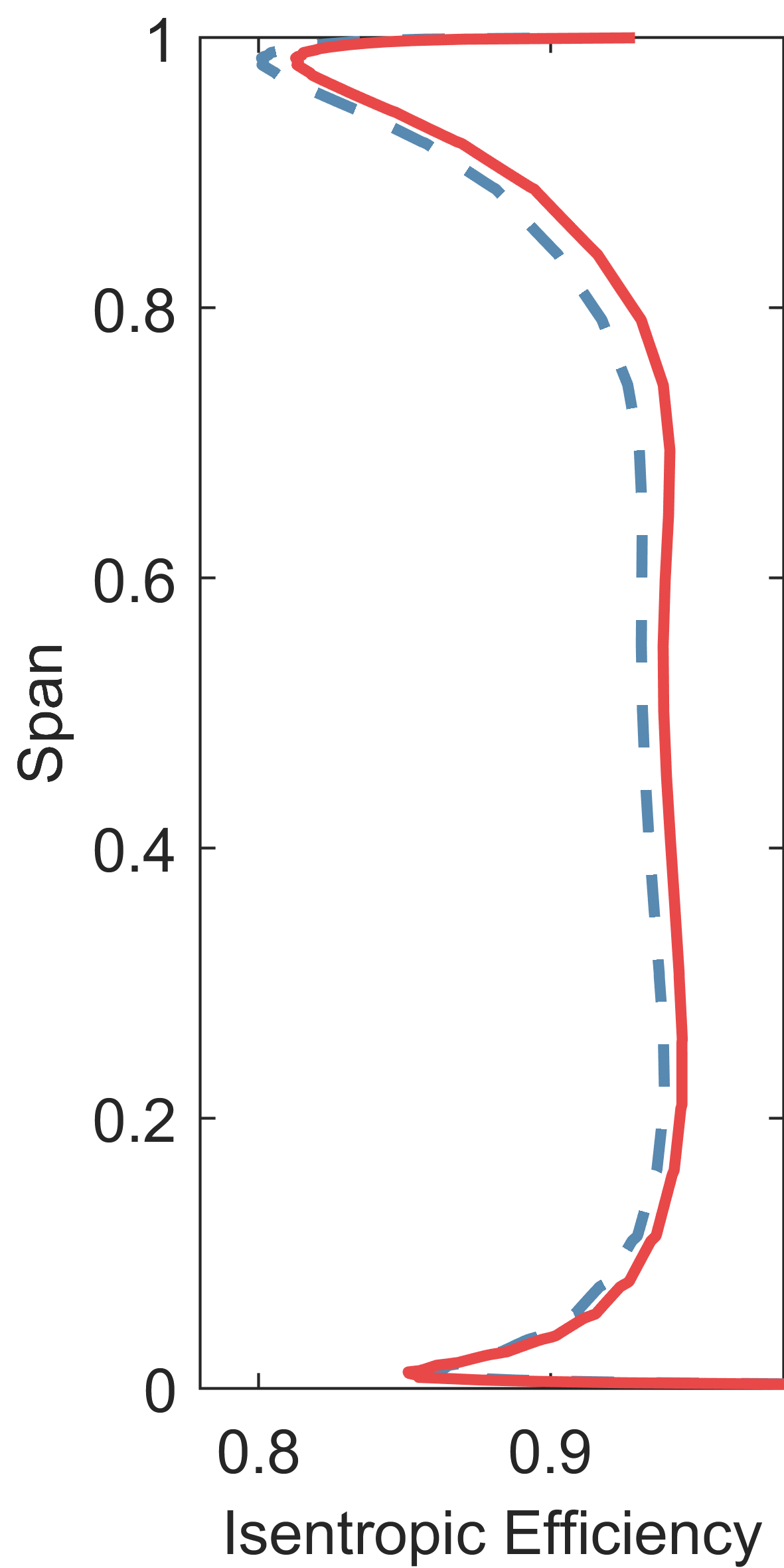}
\end{minipage}
}%
\subfigure[the second stage]{
\begin{minipage}[t]{0.33\linewidth}
\centering
\includegraphics[width=0.7\textwidth]{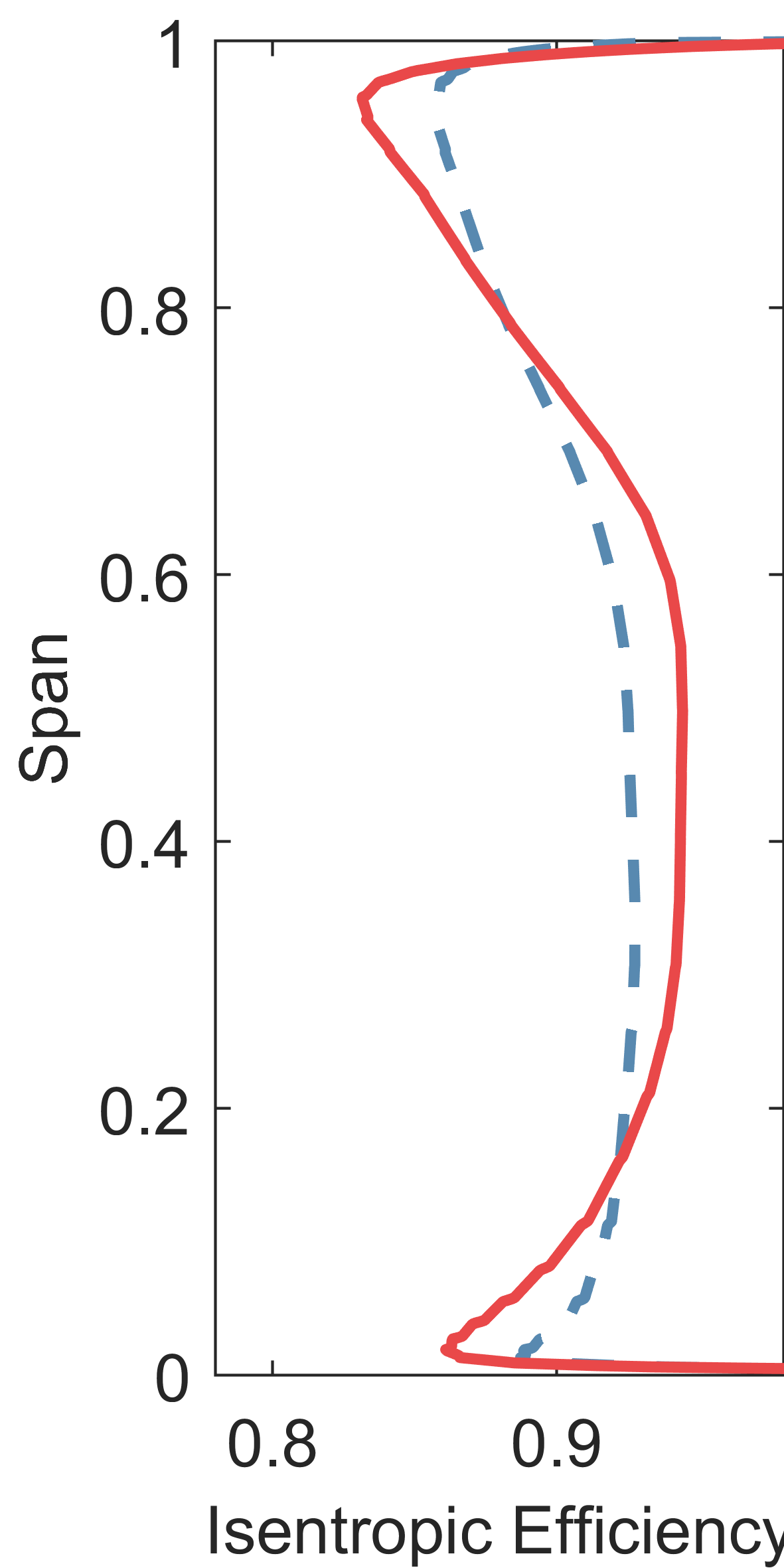}
\end{minipage}
}%
\centering
\caption{total-efficiency distribution along the blade span.}
\label{GTFeff}
\end{figure}

In Figure \ref{GTFeff}, the distribution of efficiency along the span is shown. 
The total efficiency distribution of the whole compressor, the first stage, and the second stage are shown in Fig.\ref{GTFeff}(a), (b), and (c) respectively.
Comparing the efficiency before and after optimizing by the DA-EGO algorithm: 
on the whole, the efficiency of the place higher than 90\% span changed little, 
and in the middle span increased, obviously. 
While in the place near the endwall, the efficiency decreased slightly; 
Then, in the first stage, the efficiency increased by about 1\% in all different spans; 
And in the second stage, the efficiency at the height lower than 10\% span and higher than 85\% span decreased, 
while the efficiency improvement at the middle span was very obvious.
\par
\begin{figure}[htbp]
\centering
\subfigure[baseline]{
\begin{minipage}[t]{0.9\linewidth}
\centering
\includegraphics[width=1\textwidth, trim =0cm 0cm 0cm 1.2cm,clip]{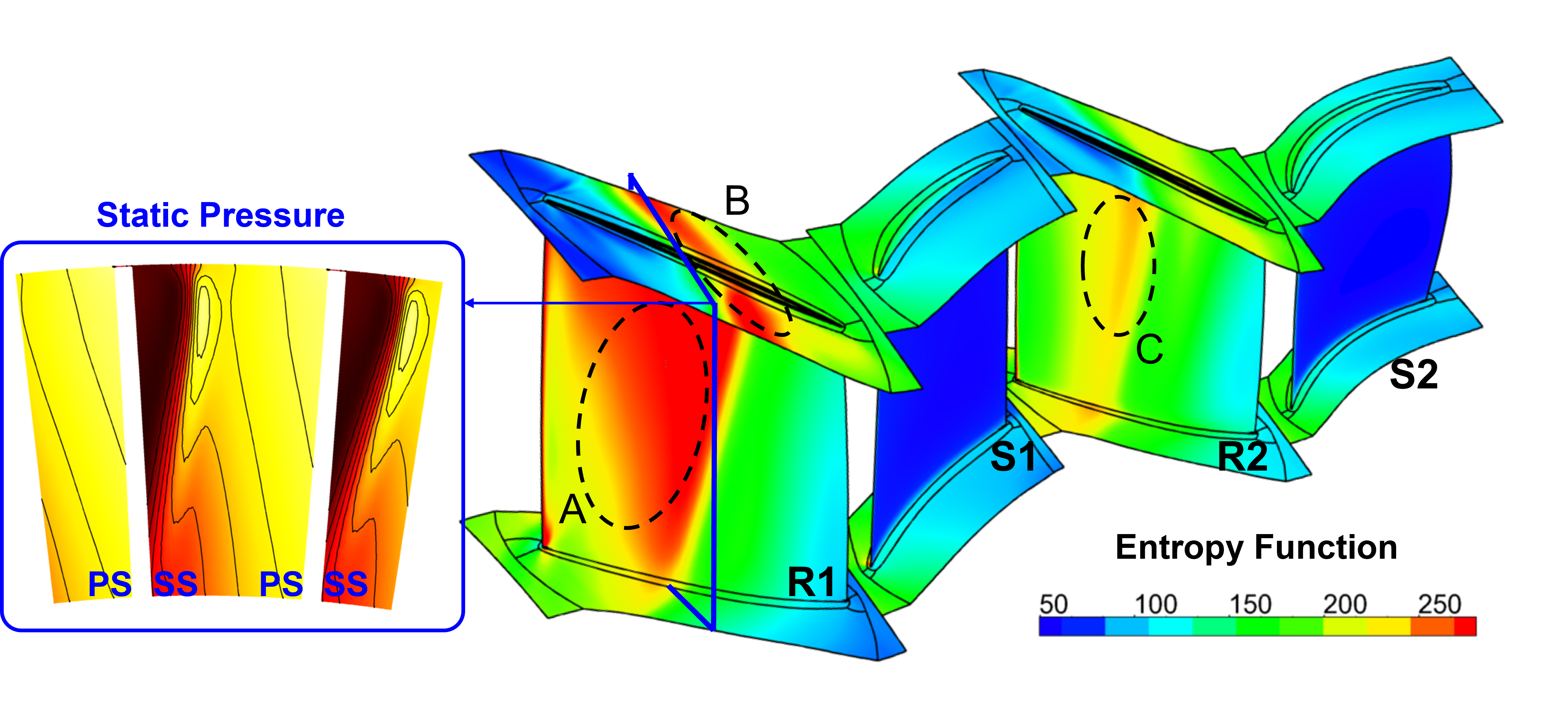}
\end{minipage}%
}%

\subfigure[optimal in DA-EGO]{
\begin{minipage}[t]{0.9\linewidth}
\centering
\includegraphics[width=1\textwidth, trim =0cm 0cm 0cm 1.2cm,clip]{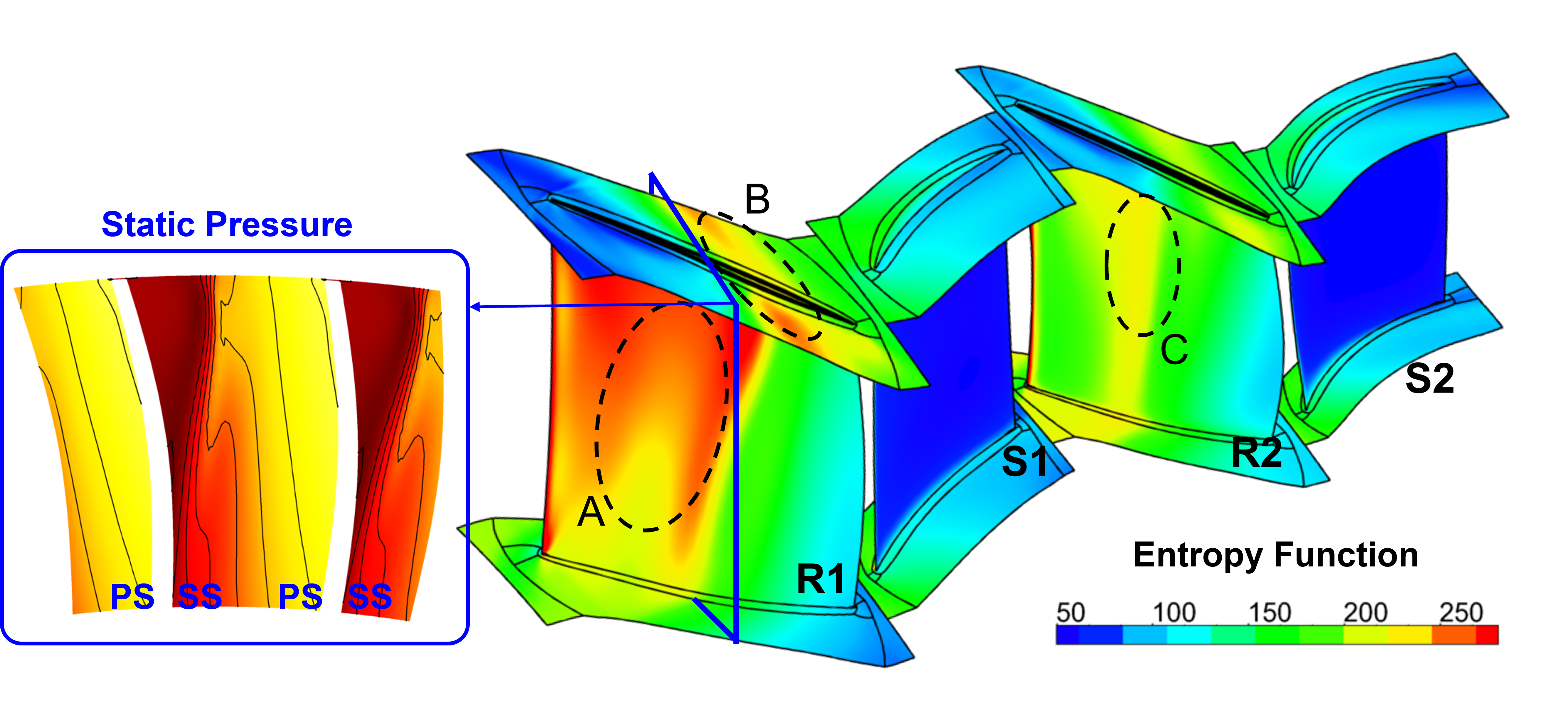}
\end{minipage}
}%
\centering
\caption{The comparison of entropy function contour of DA-EGO's optimization result and the baseline design in engineering case 2.}
\label{GTFentrop}
\end{figure}

Figure\ref{GTFentrop} shows the distribution of entropy function on the blade surface before and after optimization.
It can be seen that the three regions with the highest entropy in the flow passage (labeled as regions A, B, and C, respectively),
all show significant decreases in their entropy values after optimization. 
Among them, the entropy in regions A and B decreases most obviously. 
The reason is that the change of blade profile weakens the shock waves in rotor R1's channel, which makes the entropy around the shock waves decrease. 
Further, the static pressure distribution diagram of the middle axis section rotor R1's channel is drawn on the left side. 
After optimization, the uniformity of the circumferential static pressure distribution is greatly increased, which also proves the weakening of shock waves.  
\par
The empirical results of the engineering optimization design indicate the feasibility and practicality of the proposed DA-EGO algorithm in addressing large variable high-dimensional engineering optimization problems, with a marked improvement in efficiency as compared to other conventional methods. 
This conclusively establishes the efficacy of the DA-EGO algorithm.
Hence, the effectiveness of our proposed DA-EGO algorithm has been demonstrated.
\section{Conclusions}
\par 
In the face of regularly occurring high-dimensional engineering design challenges in real-world scenarios, this study introduces a new algorithm, the Dynamic Aggregate Efficient Global Optimization (DA-EGO) algorithm, designed specifically for resolving high-dimension and expensive black-box (HEB) issues. 
\par
The DA-EGO algorithm employs the dynamic decomposition of high-dimensional problems into multiple low-dimensional subproblems based on existing knowledge accumulated in previous cycles. 
After optimizing each subproblem, the perturbation method, and analysis of variance are used to obtain information such as the optimal point, variable interaction relationships, and so on, within each subspace. 
All knowledge obtained from subproblems is aggregated to enhance the understanding of the overall high-dimensional problem, which helps the algorithm to better decompose and set subproblems in the next cycle of optimization. 
As these steps are iterated, accumulated interaction information guides the decomposition and helps the search improve its best evaluated solution.
\par
Subsequently, this paper provides thorough validation of the DA-EGO algorithm by comparing its performance against state-of-the-art optimization algorithms across a wide range of diverse test cases. 
The test cases include 21 benchmark instances at 30, 60, and 90 dimensions, a 28-dimensional Rotor 37 blade design, and a 60-dimensional multi-stage compressor design. DA-EGO performs well on the separable and partially separable benchmarks, but GSGA is better on shifted Rosenbrock at 60 and 90 dimensions. The engineering cases support its practical value when objective evaluations are expensive.
\par
The main limitation is the additional computational cost of surrogate construction and interaction analysis. This overhead reduces the advantage of DA-EGO when objective evaluations are inexpensive. Kriging model construction is an important contributor to the algorithmic cost; more efficient Kriging implementations are a direction for further work.
\section*{Funding}
This work was supported by the National Science and Technology Major Project (2019-II-0008-0028), the Industry-University-Research Cooperation Project of Aero Engine Corporation of China (HFZL2021CXY004), and the High-level Innovative and Entrepreneurial Talents Introduction Project of Qin chuangyuan (QCYRCXM-2022-210).
\section*{Acknowledgments}
The authors would like to thank the anonymous referees for their valuable comments.
\section*{Conflict of interest}
The authors declare that they have no conflict of interest.
\section*{Data availability statement}
The data that support the findings of this study are available at \url{https://github.com/zhet1997/DA_EGO_publish}.
\bibliographystyle{elsarticle-num-names}
\bibliography{bibfile}%

\end{document}